\documentclass{article}

\usepackage[preprint, nonatbib]{neurips_2026}

\usepackage[utf8]{inputenc} % allow utf-8 input
\usepackage[T1]{fontenc}    % use 8-bit T1 fonts
\usepackage{hyperref}       % hyperlinks
\usepackage{url}            % simple URL typesetting
\usepackage{booktabs}       % professional-quality tables
\usepackage{amsfonts}       % blackboard math symbols
\usepackage{nicefrac}       % compact symbols for 1/2, etc.
\usepackage{microtype}      % microtypography
\usepackage{xcolor}         % colors
\usepackage{amsmath}
\usepackage{amssymb}
\usepackage{array}
\usepackage{tabularx}
\usepackage{booktabs}
\usepackage{xspace}
\usepackage{graphicx}
\usepackage{pifont}
\usepackage{subcaption}
\usepackage{pifont}
\usepackage{makecell}
\usepackage{multirow}
\usepackage{dsfont}
\usepackage[table, dvipsnames]{xcolor}
\usepackage[numbers]{natbib}

\definecolor{grayishred}{RGB}{250, 130, 130}
\definecolor{darkishgreen}{RGB}{35, 160, 35}
\newcommand{\cmark}{\textcolor{darkishgreen}{\ding{51}}}   % green tick
\newcommand{\xmark}{\textcolor{grayishred}{\ding{55}}}           % red cross
\newcommand{\pmark}{\textcolor{orange}{$\sim$}}           % partial / limited
\newcommand{\datasetname}{\textsc{ResCast-100K}\xspace}

\title{\datasetname: A Comprehensive Dataset for  Cross-Domain Residential Load and Indoor Temperature Forecasting}
\author{%
  Jainam Dhruva \\
  Department of Computer Science\\
  University of Kentucky\\
  Lexington, KY 40506 \\
  \texttt{jainam.dhruva@uky.edu} \\
  \And
  Yousaf Raza \\
  Department of Computer Science\\
  University of Kentucky\\
  Lexington, KY 40506 \\
  \texttt{yra237@uky.edu} \\
  \AND
  A.B Siddique \\
  Department of Computer Science\\
  University of Kentucky\\
  Lexington, KY 40506 \\
  \texttt{msi290@g.uky.edu} \\
  \And
  Simone Silvestri \\
  Department of Computer Science\\
  University of Kentucky\\
  Lexington, KY 40506 \\
  \texttt{simone.silvestri@uky.edu} \\
}

\begin{document}

\maketitle

\begin{abstract}
Accurate short-term forecasting of residential energy load and indoor temperature is essential for home energy management systems, grid-level demand response participation, and community-level energy efficiency programs.
Domain adaptation methods and transfer learning have shown promise in improving forecasting accuracy under data heterogeneity and scarcity scenarios commonly observed at the residential level.
However, progress remains limited by the absence of comprehensive residential datasets: existing ones are narrow in target coverage and lack structured support for cross-domain evaluation.
%However, progress remains limited due to the absence of large-scale residential datasets that support systematic cross-domain evaluations.
% Recent works do not permit domain construction along different interpretable axes — such as geography, climate zone, wall construction, heating equipment, and more — lacking benchmark support for zero-shot, transfer-learning, and domain adaptation methodologies. 
%
To address this, we introduce {\datasetname}, a large-scale residential forecasting benchmark for studying cross-domain generalization. It exposes a configuration-driven interface that instantiates source and target domains along interpretable axes, such as geography, climate zone, wall construction, and heating equipment, supporting systematic evaluation of transfer learning, domain adaptation, and zero-shot generalization under controlled domain shifts. 
The benchmark covers approximately 100,000 EnergyPlus-simulated U.S. homes derived from ResStock, providing 15-minute time series of three coupled targets per home, namely total load, HVAC (heating, ventilation, and air conditioning) load, and indoor temperature, together with weather channels, HVAC setpoints, and over 40 static building covariates. 
Finally, it integrates five real-world residential datasets under a unified schema, supporting sim-to-real evaluation on the same tasks.
%
% To address this, we introduce {\datasetname}, a large-scale residential forecasting benchmark of 15-minute time series for approximately 100,000 simulated U.S. homes. {\datasetname} includes total load, heating, ventilation, and air-conditioning~(HVAC) load, and indoor temperature data with weather channels, HVAC setpoints, and over 40 static building co-variates.
% The dataset is curated through high-fidelity \textit{EnergyPlus} simulations. The benchmark exposes a configuration-driven interface that aids instantiating domain splits along interpretable axes - such as geography, climate zone, wall construction, heating equipment, and more - supporting domain adaptation methodologies' evaluations.
% We further integrate five real-world residential datasets under a unified schema, to support sim-to-real evaluations. 
We benchmark models from recurrent, attention, and MLP-mixer architectures for zero-shot performance across different domains, under missing input data conditions, for each task. 
Cross-attention and MLP-mixer based models consistently outperform recurrent and classical transformer-based baselines under domain shift. 
{\datasetname} is intended to support the 
machine learning and building analytics communities in advancing cross-domain residential forecasting at the home, community, and grid levels.
%It is also observed that probabilistic forecasts trained on synthetic data transfer more reliably to real homes than point estimates. 
%Overall, this benchmark aims to serve both machine learning and building analytics communities in developing cross-domain forecasting methodologies, and evaluating their impact on achieving energy goals at the home, community, and grid levels.
\end{abstract}
\section{Introduction}
\label{sec: Introduction}
Accurate short-term forecasting of residential energy load and indoor temperature is critical for modern data-driven energy systems, enabling applications such as intelligent home energy management, grid balancing, and demand response~\cite{multizoneIndoorTemp, transfer_for_demand_transformer, dhruva_hvac_control, xu_transfer_hvac}. 
Recent advances in deep learning have improved forecasting accuracy substantially, but these models typically require access to large amounts of historical training data from the target domain~\cite{re_alden, timexer, patchTST}. 
Residential settings rarely meet this requirement: data is often scarce, fragmented, or privacy-sensitive, motivating forecasting models that generalize across domains.

% While recent advances in deep learning-based models have significantly improved forecasting performance, their success typically relies on access to large amounts of historical data from the target domain~\cite{re_alden, timexer, patchTST}. 
% However, in residential settings, such data is often scarce, fragmented, or privacy-sensitive, motivating the need to utilize forecasting models that can generalize across domains.

% making it difficult to deploy robust forecasting models at scale.

% Hence, energy management systems need to utilize forecasting models that can generalize across domains, where models trained on one set of buildings transfer to others with different characteristics. 

% However, 

Residential environments exhibit multi-dimensional heterogeneity across geographic location, climate, building envelope, and HVAC systems, where each axis induces a distinct distribution shift for total load, HVAC load, and indoor temperature forecasting. 
Domain adaptation and transfer learning have emerged as promising approaches under such heterogeneity~\cite{buildingsBench, tl_and_xai_for_load, multizoneIndoorTemp}.
Despite this growing interest, progress remains hindered by the lack of large-scale datasets that support systematic evaluation under controlled domain shifts. 
Existing datasets are limited in one or more aspects: they focus on non-residential buildings, omit key exogenous variables such as weather or control signals, lack targets such as HVAC load or indoor temperature, or do not span enough diversity to study generalization~\cite{buildingsBench, hotDataset, btsDataset, bdg2}. 
Moreover, prior benchmarks rarely integrate synthetic and real-world 
data under a unified schema, making sim-to-real evaluation difficult.

\begin{table*}[h]
\centering
\caption{Comparison of \textsc{ResCast-100K} with existing building energy and temperature datasets. Categories are grouped into \emph{Scope}, \emph{Targets}, \emph{Covariates}, and \emph{Cross-Domain Support}. \cmark{} = supported, \xmark{}~=~not supported, and \pmark{}~=~partial~/~limited. ``Resi.''~=~residential.}
\label{tab:dataset_comparison_qualitative}
\setlength{\tabcolsep}{3pt}
\renewcommand{\arraystretch}{1.05}
\scriptsize
\begin{tabular}{l cc ccc ccc ccccc}
\toprule
& \multicolumn{2}{c}{\textbf{Scope}}
& \multicolumn{3}{c}{\textbf{Targets}}
& \multicolumn{3}{c}{\textbf{Covariates}}
& \multicolumn{5}{c}{\textbf{Cross-Domain Support}} \\
\cmidrule(lr){2-3} \cmidrule(lr){4-6} \cmidrule(lr){7-9} \cmidrule(lr){10-14}
\textbf{Dataset}
& Resi. & Open
& Total & HVAC  & Indoor
& Weather & Control & Static
& Geo. & Climate & Envel. & HVAC & Sim+ \\
& focus & access
& load  & load  & temp.
& channels & setpoints & covs.
& split & split & split & split & Real \\
\midrule
UCI Electricity~\cite{uci_elec_load_diagrams}
& \pmark & \cmark & \cmark & \xmark & \xmark & \xmark & \xmark & \xmark & \xmark & \xmark & \xmark & \xmark & \xmark \\
Ind.\ Household Power~\cite{uci_individual_household}
& \cmark & \cmark & \cmark & \xmark & \xmark & \xmark & \xmark & \xmark & \xmark & \xmark & \xmark & \xmark & \xmark \\
Low Carbon London~\cite{low_carbon_london}
& \cmark & \cmark & \cmark & \xmark & \xmark & \xmark & \xmark & \xmark & \xmark & \xmark & \xmark & \xmark & \xmark \\
Pecan Street~\cite{pecan_street}
& \cmark & \xmark & \cmark & \pmark & \xmark & \xmark & \xmark & \pmark & \cmark & \pmark & \xmark & \xmark & \xmark \\
Ecobee DYD~\cite{ecobee_dataset}
& \cmark & \xmark & \xmark & \xmark & \cmark & \pmark & \cmark & \pmark & \pmark & \pmark & \xmark & \xmark & \xmark \\
BDG2~\cite{bdg2}
& \xmark & \cmark & \cmark & \xmark & \xmark & \cmark & \xmark & \pmark & \cmark & \cmark & \xmark & \xmark & \xmark \\
LBNL Bldg.\ 59~\cite{lbnl59}
& \xmark & \cmark & \cmark & \cmark & \cmark & \cmark & \cmark & \pmark & \xmark & \xmark & \xmark & \xmark & \xmark \\
BTS~\cite{btsDataset}
& \xmark & \cmark & \cmark & \pmark & \cmark & \cmark & \cmark & \pmark & \xmark & \xmark & \xmark & \xmark & \xmark \\
BuildingsBench~\cite{buildingsBench}
& \pmark & \cmark & \cmark & \xmark & \xmark & \pmark & \xmark & \pmark & \cmark & \pmark & \xmark & \xmark & \cmark \\
\midrule
\rowcolor{green!8}
\textbf{\textsc{ResCast-100K} (Ours)}
& \cmark & \cmark & \cmark & \cmark & \cmark & \cmark & \cmark & \cmark & \cmark & \cmark & \cmark & \cmark & \cmark \\
\bottomrule
\end{tabular}
\end{table*}

To address these limitations, we introduce \datasetname, a large-scale dataset and benchmark for cross-domain residential forecasting. The dataset contains 15-minute resolution time series for over 100{,}000 homes across the United States, including total load, HVAC load, and indoor temperature, along with rich exogenous variables such as weather and control inputs, and static building covariates. It integrates high-fidelity physics-based simulations with curated real-world datasets under a unified schema, enabling both controlled experimentation and realistic evaluation scenarios. 

A key feature of \datasetname\ is its configuration-driven domain abstraction, which allows users to define domain splits along multiple interpretable axes, including but not limited to, geography, climate zone, construction type, and HVAC technology. This facilitates systematic evaluation of timely methods within transfer learning, meta-learning, domain adaptation, and zero-shot generalization under diverse distribution shifts. Additionally, the combination of synthetic and real data enables study of synthetic-to-real transfer, which is essential for scaling methods in data-constrained environments.

Using this benchmark, we evaluate commonly used and state of the art time-series models, including recurrent, transformer-based, and mixer-style architectures, under both standard and missing-covariate settings. Our results demonstrate that cross-attention and MLP-mixer based models retain best accuracy under domain shift across all three forecasting tasks. These two architectures also achieve the best sim-to-real probabilistic forecasting accuracy.
% We also observe that probabilistic forecasting models trained on synthetic data transfer more reliably to real-world settings than point prediction models. 

Overall, this work makes the following contributions:
\begin{itemize}
    \item We introduce \datasetname, a large-scale residential dataset of 100K homes with a configuration-driven interface that enables controlled study of cross-domain forecasting along axes such as geography, climate zone, construction type, and heating equipment.
    \item The dataset provides multiple coupled targets — total load, HVAC load, and indoor temperature — essential for downstream tasks, together with exogenous features including weather variables, HVAC setpoints, and static covariates characterizing each home.
    \item We benchmark recurrent, attention, and MLP-mixer style models and provide insights for model-architecture choices. We further integrate synthetic and real-world datasets under a unified schema and provide zero shot sim-to-real evaluations.
\end{itemize}

% \begin{figure}[h]
% \centering
% \includegraphics[width=\linewidth]{Figures/KDE.png}
% \caption{KDE}
% \label{fig:kde}
% \end{figure}

\section{Related Work}
\label{sec: related_works}
% Short-term forecasting of the load and indoor temperature has become an an essential utility for modern energy systems for supporting grid balancing, reducing energy bills, optimal renewable utilization, demand response, and occupant-aware building-level and grid level energy management systems \citep{zhang2021review, buildingsBench, transfer_for_demand_transformer, dhruva_hvac_control}. Accurate forecasts are frequently used as inputs in predictive control, reinforcement-learning-based control, scheduling, and home energy management systems, where forecast errors propagate directly into energy cost, comfort violations, and flexibility estimates \citep{drgovna2020all, killian2016ten, zhang2021review}. Overall, residential sector accounts for roughly two-thirds of the building energy usage globally~\cite{review_on_buildings}, and hence is a sector with considerable promise for energy conservation and flexibility, which to be realized, requires accurate forecasting.

\textbf{Current Datasets.} Public and open source datasets of building operational data have been scarce, narrow in scope, leave out on information-rich co variates, or are often inaccessible for reproducible research. Aggregate consumption datasets such as the UCI Electricity Load Diagrams~\cite{uci_elec_load_diagrams}, Low Carbon London dataset~\cite{low_carbon_london}, and the Individual Household Electric Power Consumption dataset~\cite{uci_individual_household}, common in machine learning and time series benchmarking, provide hourly or sub-hourly load measurements; but they are restricted to a single channel of total electricity demand without coupled weather, control, or information about the homes. Some larger datasets such as the Building Data Genome Project 2 (BDG2)~\cite{bdg2, bdg1} aggregate 3053 meters across 1636 buildings, while the Building Performance Database~\cite{building_performance_db} catalogs over a million buildings; however, both remain dominated by commercial stock, rely on coarse temporal resolutions, and/or also do not include relevant covariates. BuildingsBench~\cite{buildingsBench} releases Buildings-900K, a simulated corpus of buildings derived from Resstock for enabling pretraining of energy forecasting models. But this dataset lacks granular information, does not capture exogenous features, lacks HVAC load and indoor temperature target series essential for energy management systems, and does not support systematic cross domain adaptation. On the contrary, some holistic datasets such as Lawrence Berkeley National Laboratory's Building 59 (LBNL59)~\cite{lbnl59} and the Building TimeSeries (BTS) dataset~\cite{btsDataset} expose comprehensive operational signals---including HVAC states, setpoints, and zone temperatures---but represent only one and three buildings, respectively, severely limiting their utility for studying generalization across domains. 
Other resources, such as Pecan Street~\cite{pecan_street}, are either gated, exclude important exogenous features, or are restricted to specific geographies. 
% Recent datasets do no support cross domain generalization a do no include the three targets.

\textbf{Scaling Through Simulations.}
To overcome the scaling limits of real-world data, recent work has turned to high-fidelity physics-based simulation. The U.S. Department of Energy's ResStock~\cite{resstock} and ComStock~\cite{ComStock} provide statistically representative models of U.S. residential and commercial stock, calibrated against advanced metering data and surveys such as CBECS~\cite{CBECS}. Models pre-trained on synthetic corpora generated by simulators such as EnergyPlus have shown preliminary success in generalizing to real buildings in both zero-shot and fine-tuned settings~\cite{buildingsBench,transfer-of-sim2real, physics-adaptation, sim-2-real}, and sim-to-real transfer is helpful in the residential setting, where smart-meter and smart-thermostat data are scarce due to collection cost and privacy concerns~\cite{us_energy_profiles, nilm_dataset}. But most efforts predominantly target commercial buildings, and sim-to-real transfer for residential households remains underexplored, due to higher heterogeneity and limited structured real-world data~\cite{buildingsBench,us_energy_profiles}. 
% Therefore, large-scale synthetic datasets coupled with structured real datasets are needed to scale residential forecasting beyond what real-world data alone can support.

\textbf{Forecasting in Residential Setting.} 
% At the building scale, in residential setting, the forecasting problem is particularly challenging: household loads are strongly influenced by stochastic occupant behavior, appliance usage, weather, envelope properties, equipment type, and thermostat operation~\cite{hvac_transfer, casella_hvac_conservation, mulayim2024time,li2022building}. 
% Unlike many commercial buildings, homes often exhibit sparse sensing, heterogeneous metadata, irregular control inputs, and high inter-household variability. 
Several works have focused on developing better models for improving forecasting within residential domain. The problem has been shown to be particularly challenging as household loads are strongly influenced by stochastic occupant behavior, appliance usage, weather, envelope properties, equipment type, and thermostat operation~\cite{hvac_transfer, casella_hvac_conservation, mulayim2024time,li2022building}. 
Previous works have benchmarked transformer style and Mixer style model architectures for time series forecasting such as Informer~\cite{zhou2021informer}, PatchTST~\cite{patchTST}, TSMixer~\cite{TSMixer}, and TimeXer~\cite{timexer}. However, most evaluations focus on univariate-input based forecasting, focus on non-residential setting, and primarily lack cross domain generalization benchmarking of these approaches~\cite{genTL, buildingsBench, Real_E}. Models trained on one population of homes may fail under shifts in geography, climate zone, construction type, heating and cooling equipment, occupancy pattern, or data availability~\cite{xing2024transfer, genTL, hvac_transfer, tl_and_xai_for_load}. 
% These characteristics make residential forecasting not only a time-series prediction problem, but also a domain generalization problem. Hence, there is a need of forecasting models, and with that the need for datasets that empower them, so that that these forecasting models can train and adapt quickly to different systems. 

\begin{figure}[h]
\centering
% \fbox{\includegraphics[width=0.95\linewidth,height=0.125\textheight]{Figures/dynamic_feature_distribution_overlays.pdf}}
\fbox{\includegraphics[width=0.95\linewidth,height=0.125\textheight]{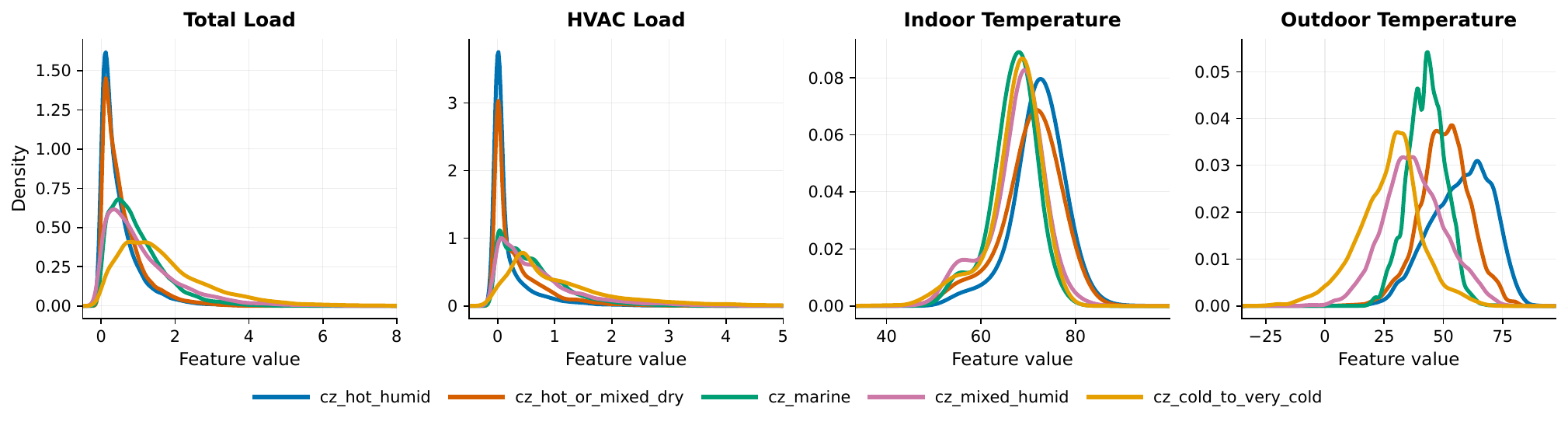}}
\caption{Distribution of total load (kWh), HVAC load (kWh), indoor temperature (°F), and outdoor temperature (°F) across the five ASHRAE/IECC climate-zone domains: Hot-Humid, Hot/Mixed-Dry, Marine, Mixed-Humid and Cold/Very-Cold. Distributions are estimated via kernel density across all homes within each climate-zone domain. The observed shifts in target-variable distributions indicate cross-domain heterogeneity. When coupled with exogenous covariates and temporal dependencies, these shifts further increase across the domains.}
\label{fig:distribution_shift}
\end{figure}

\textbf{Cross-Domain Generalization in Residential Buildings.}
Transfer learning has emerged as a principal approach to scalable deployment of data-driven controllers and forecasters across buildings, given the prohibitive cost of per-building data collection and customization. Reinforcement learning based methods in HVAC-control studies include Xu et al.~\cite{xu_transfer_hvac}, 
% who transfer DQN policies across multi-zone buildings with only two weeks of adaptation data, 
Coraci et al.~\cite{coraci_transfer}, 
% who report 50--80\% reductions in temperature violations through online Soft Actor-Critic transfer, 
and Lissa et al.~\cite{lissa_transfer}, who transfer control policies across buildings, room, and microgrid-level heat-pump control. 
However, model based approaches remain preferred in the building community due to their inherent nature of being more explainable. Model based methods have investigated meta-learning approaches such as MAML~\cite{finn2017model} and its forecasting extensions~\cite{he_maml}. 
Some other work have evaluated target domain adaptation for residential total load, HVAC load, and indoor temperature forecasting, using Graph Neural Networks~\cite{gnn_cite}, Long Short Term Memory models~\cite{lstm_cite}, and Transformers~\cite{attentionIsAllYouNeed}, under different source and target set cardinalities, but on smaller, individual, and/or non-public datasets~\cite{gnn_total_transfer, hvac_transfer, therml_comfort_transfer, genTL}. 
% BuildingsBench~\cite{buildingsBench} demonstrated synthetic pretraining yields surprisingly strong generalization that to real commercial buildings while leaving substantial gaps in residential buildings and leaving it as an open problem. 
These works focus on smaller domain perturbations under individual and/or non-public dataset, making it difficult to compare the cross-domain generalization efficacy of those approaches. 
% This has today necessitated the need for public and open source residential dataset enables systematic transfer evaluation under structured shift.

\textbf{\datasetname.} \datasetname is purpose-built to close gaps observed in recent literature. This comprehensive dataset multi-target timeseries data for approximately 100{,}000 homes across all U.S. states with paired weather, setpoints, and metadata; integrates five real residential datasets under a consistent schema; and exposes a configuration-based interface for instantiating cross-domain splits. Hence, \datasetname enables transfer-learning, domain-adaptation, zero-shot, and sim-to-real evaluations that prior residential forecasting resources could not support.

\section{Dataset Description}
\label{sec: dataset_description}
\paragraph{\textbf{Overview.}} We introduce the \datasetname dataset, for cross domain forecasting in residential setting. Our dataset provides a 15-minute granular view where total load, HVAC load, and the indoor temperature data are coupled with exogenous features and static covariates (refer section \ref{appendix:static_covariates}). The exogenous features allow to better capture the patterns in the target. The static variables provide information that can be used for generalization by the model, as well as to evaluate the transferability proxy from one building to another. 

\begin{figure}[h]
\centering
\fbox{\includegraphics[width=\linewidth]{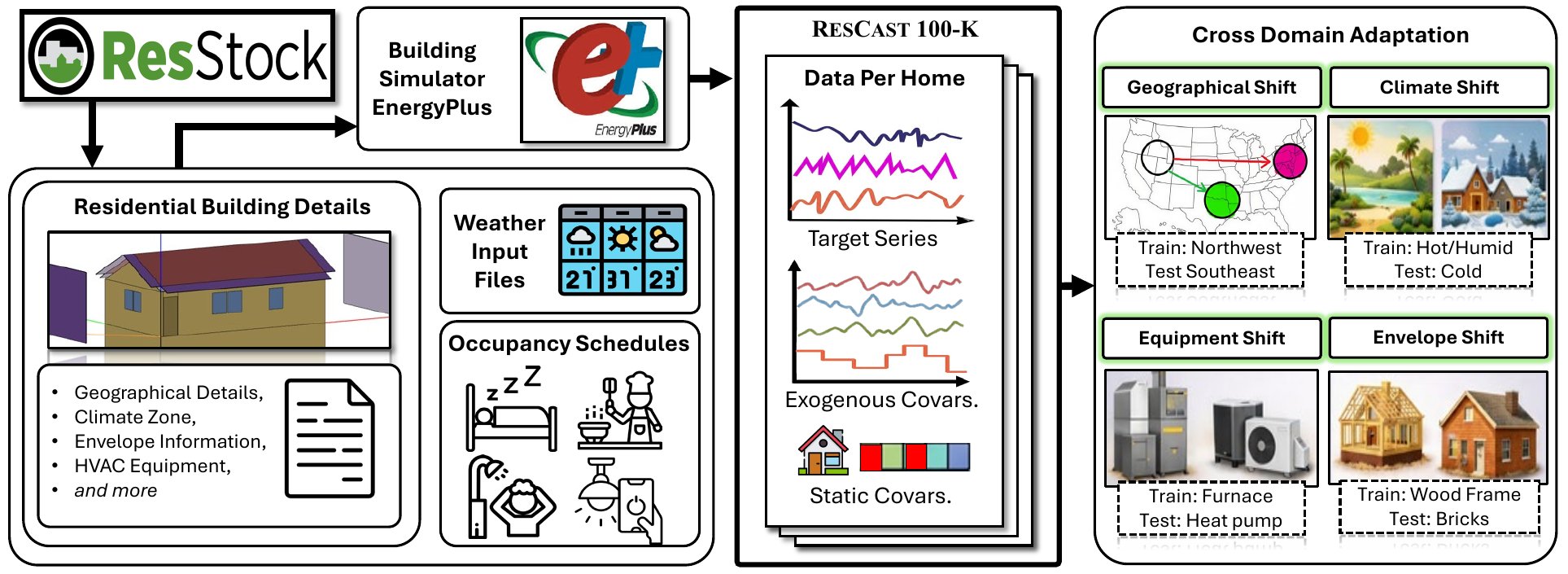}}
\caption{Data Curation pipeline for \datasetname. Home envelope files, weather data, and stochastic schedules, derived from Resstock are used as inputs to EnergyPlus simulations computed through high performance computing (HPC) cluster.  The data can be used to create specific domain splits for cross domain adaptions.
% , as well as for synthetic and real data split to evaluate synthetic to real adaptation. 
This dataset can be used to evaluate various domain adaptation methodologies for time series forecasting in residential setting.}
\label{fig:data_curation}
\end{figure}

\paragraph{\textbf{Curation.}} We construct the dataset with a pipeline built on ResStock and EnergyPlus, executed on high-performance computing nodes. ResStock~\cite{resstock} provides a statistically representative distribution of U.S. housing — envelope types, regions, occupancy schedules, and equipment — calibrated against research studies, census records, and surveys~\cite{CBECS, stochasticSchedule}. This is paired with approximately 9,000 TMY3 (typical meteorological year) weather files covering the United States. The homes are then conditionally sampled together with their corresponding weather files to produce simulation inputs for roughly 100,000 buildings across the United States.
To prevent under-representation of demographically smaller or data-scarce regions, we enforce a near-uniform allocation of approximately 2,000 homes per state. 
% We then simulate one year of operation per home in EnergyPlus\cite{energyplus}, the U.S. Department of Energy's reference whole-building simulator, to obtain the target time series. 
To obtain the target time series, we simulate one year of operation per home in EnergyPlus~\cite{energyplus}, the U.S. Department of Energy's reference whole-building simulator. EnergyPlus is the industry standard for building energy modeling, with validation against the ASHRAE Standard 140 method of test for building energy simulation programs~\cite{ashrae140}.
% This uses over $100$ compute hours for the final simulation. 

\paragraph{\textbf{Features and structure.}} The dataset is sharded into a set of Parquet files, with each consisting of yearly time series data for a batch of homes. The Parquet file contains building ID, timestamps, the three targets (total load, HVAC load, indoor temperature), and the exogenous features. Home properties, i.e., the static covariates — geographical attributes, envelope properties, HVAC system descriptors, etc. — are consolidated into a single parquet file indexed by building ID. To enable sim-to-real evaluation, we include five real residential datasets, summarized in Table \ref{tab:real_dataset_overview}. Real datasets vary in coverage: some lack weather channels, others omit setpoints, and most expose only a small subset of static home properties. We structure them under a common schema through a combination of automated pre-processing — resampling, outlier removal, and missing-value imputation — and manual intervention to align static-variable definitions across sources.

\begin{table}[h]
\centering
\caption{Summary of time series data for each residential dwelling: exogenous covariates, static covariates, and target variables. Files also include respective timesteps for that time series data.}
\label{tab:features}
\begin{tabular}{p{0.42\linewidth} p{0.42\linewidth}}
  \toprule
  \textbf{Exogenous features} & \textbf{Static Covariates and Target Features} \\
  \midrule

  Outdoor Drybulb temp. ($^\circ$F) \newline
  Outdoor Wetbulb temp. ($^\circ$F) \newline
  Relative humidity (\%) \newline
  Wind speed (m/s) \newline
  Diffuse solar radiation (W/m$^2$) \newline
  Direct solar radiation (W/m$^2$) \newline
  Heating setpoint ($^\circ$F) \newline
  Cooling setpoint ($^\circ$F)
  &
  \begin{tabular}[t]{@{}p{\linewidth}@{}}
    \textbf{Static properties:} \newline
    State, Climate zone, Floor area,
    Wall type, Foundation type, HVAC system type, etc. \\
    \midrule
    \textbf{Targets:} \newline
    Total electric load (kWh) \newline
    HVAC electric load (kWh) \newline
    Indoor temperature ($^\circ$F)
  \end{tabular}
  \\

  \bottomrule
\end{tabular}
\end{table}

\paragraph{\textbf{Usage.}} The dataset code provides a configuration-driven data loader that instantiates a domain from a user-defined YAML file. Users specify a domain by listing values along any axis exposed in the metadata: states, cities, climate zones, floor-area ranges, wall types, heating equipment, and so on. We provide a dictionary of domain axes and the categorical values it can consist of. The code provides a Pytorch style dataset and dataloader for source and target domain as defined in the YAML config file. The dataloader handles both synthetic and real datasets and returns PyTorch-compatible iterators, so any user-defined or pre-defined model can be trained and evaluated under identical interfaces. The data curation pipeline and its application are summarized in Figure \ref{fig:data_curation}.

\begin{table*}[h]
\centering
\caption{Overview of real datasets included in \datasetname.}
\small
\begin{tabularx}{\textwidth}{
l
>{\centering\arraybackslash}p{0.6cm}
>{\centering\arraybackslash}p{0.6cm}
>{\centering\arraybackslash}p{0.6cm}
c
>{\centering\arraybackslash}p{0.7cm}
>{\raggedright\arraybackslash}X
>{\raggedright\arraybackslash}X
}
\toprule
Dataset & Indoor temp. & HVAC Load & Total Load & Homes & Static Vars & Geo Locations Represented & Data Time Range \\
\midrule
ECOBEE\cite{ecobee_dataset} & \checkmark & $\times$ & $\times$ & 954 & 23 & $\approx 43$ cities & Jan 2017 -- Dec 2017 \\
HEAPO\cite{heapo_dataset}  & $\times$ & $\times$ & \checkmark & 1,407 & 9 & $\approx 1$ city & May 2019 -- Feb 2024 \\
IDEAL\cite{ideal_dataset}  & \checkmark & $\times$ & \checkmark & 255 & 19 & $\approx 5$ & Aug 2016 -- Jun 2018 \\
NEST\cite{nest_dataset}  & \checkmark & \checkmark & \checkmark & 1 & 20 & $\approx 1$ city & Jul 2022 -- Jun 2023 \\
REFIT\cite{refit_dataset}  & $\times$ & $\times$ & \checkmark & 20 & 11 & Not available & Sep 2013 -- Jul 2015 \\
\bottomrule
\end{tabularx}
\label{tab:real_dataset_overview}
\end{table*}
\section{Problem Statement}
\label{sec: problem_statement}
RESCAST-100K supports a broad class of residential forecasting problems, including supervised learning, transfer learning, domain adaptation, zero-shot generalization, robustness to missing covariates, and sim-to-real evaluation. Here, we instantiate one representative setting: probabilistic forecasting under domain transfer, with the goal of minimizing expected forecasting loss across shifted domains while handling missing dynamic and static covariates.
\paragraph{Forecasting setup.}
We formulate residential forecasting as a univariate forecasting task instantiated independently for each of the three targets (Total, HVAC, Temp.), with multivariate dynamic and static inputs. Let $L_c$ and $L_f$ denote context length and forecast horizon. For each sample $i$,
\[
\mathbf{x}^{(i)}_{1:L_c} \in \mathbb{R}^{N_p \times L_c},\quad
\mathbf{u}^{(i)}_{1:L_f} \in \mathbb{R}^{N_f \times L_f},\quad
\mathbf{s}^{(i)} \in \mathbb{R}^{N_s},\quad
\mathbf{y}^{(i)}_{1:L_f} \in \mathbb{R}^{1 \times L_f},
\]
where $\mathbf{x}_{1:L_c}$ stacks the past target and past covariates ($N_p$ channels), $\mathbf{u}_{1:L_f}$ contains future-known dynamic inputs ($N_f$ channels, e.g., weather forecasts and scheduled setpoints), $\mathbf{s}$ contains static home attributes ($N_s$ entries), and $\mathbf{y}_{1:L_f}$ is the target trajectory. We learn a quantile predictor
\begin{equation}
\label{eq:predictor}
    f_\theta : (\mathbf{x}_{1:L_c},\, \mathbf{u}_{1:L_f},\, \mathbf{s}) \mapsto \hat{\mathbf{Q}}_\theta \in \mathbb{R}^{L_f \times Q},
\end{equation}
where $\hat{Q}_{t,\tau_q}$ estimates the $\tau_q$-quantile of $Y_t$ for levels $\tau_1<\dots<\tau_Q \in (0,1)$, trained by minimizing the average pinball loss $\mathcal{L}_{\mathrm{prob}}$ over the horizon and quantile grid (Appendix~\ref{app:loss}):
\begin{equation}
    \min_\theta\; \mathbb{E}_{(\mathbf{x},\mathbf{u},\mathbf{s},\mathbf{y})\sim P}\bigl[\mathcal{L}_{\mathrm{prob}}\bigl(f_\theta(\mathbf{x},\mathbf{u},\mathbf{s}),\,\mathbf{y}\bigr)\bigr].
\end{equation}
Probabilistic forecasts are essential in residential settings, where downstream control and planning rely on uncertainty quantification~\cite{transfer_for_demand_transformer, prob_tft, clue_rl}.

\paragraph{Cross-domain generalization.}
Homes are drawn from heterogeneous distributions induced by geography, climate, envelope, and equipment. Each domain $d \in \mathcal{D}$ can be modeled with joint $P_d(\mathbf{x}_{1:L_c}, \mathbf{u}_{1:L_f}, \mathbf{s}, \mathbf{y}_{1:L_f})$, defined by fixing values along a designated subset of static attributes (e.g., climate zone, wall type); the remaining static attributes vary within each domain and are passed to $f_\theta$. Given source domains $\mathcal{D}_{\mathrm{src}} = \{d_1,\dots,d_K\}$, training minimizes the expected risk under a meta-distribution $\Pi_{\mathrm{src}}$ corresponding to sample-weighted pooling of source-domain data:
\begin{equation}
    \min_\theta\; \mathbb{E}_{d \sim \Pi_{\mathrm{src}}}\,\mathbb{E}_{(\mathbf{x},\mathbf{u},\mathbf{s},\mathbf{y})\sim P_d}\bigl[\mathcal{L}_{\mathrm{prob}}\bigl(f_\theta(\mathbf{x},\mathbf{u},\mathbf{s}),\,\mathbf{y}\bigr)\bigr].
\end{equation}
Generalization is assessed zero-shot on a held-out target domain $d^\star \notin \mathcal{D}_{\mathrm{src}}$, or with limited adaptation data in the transfer-learning setting. Domain splits are instantiated by the user via a YAML config provided with the dataset.

\paragraph{Missing covariates.}
In practice, weather channels, setpoints, or home metadata may be systematically absent due to lack of sensing, older thermostat and metering infrastructure, incomplete metadata collection, or inconsistencies across data providers. We model this via a stochastic masking operator $\mathcal{M} \sim \mathcal{P}_M$ that drops complete dynamic channels independently and the static-covariate vector jointly, yielding $(\tilde{\mathbf{x}}, \tilde{\mathbf{u}}, \tilde{\mathbf{s}}) = \mathcal{M}(\mathbf{x}, \mathbf{u}, \mathbf{s})$; and the target is never masked. The full training objective is
\begin{equation}
    \min_\theta\; \mathbb{E}_{d \sim \Pi_{\mathrm{src}}}\,\mathbb{E}_{(\mathbf{x},\mathbf{u},\mathbf{s},\mathbf{y})\sim P_d}\,\mathbb{E}_{\mathcal{M}\sim\mathcal{P}_M}\bigl[\mathcal{L}_{\mathrm{prob}}\bigl(f_\theta(\tilde{\mathbf{x}}, \tilde{\mathbf{u}}, \tilde{\mathbf{s}}),\,\mathbf{y}\bigr)\bigr].
\end{equation}
Thus, our formulation seeks to learn a probabilistic forecaster that generalizes across structured domain shifts while remaining robust to incomplete inputs.
\section{Experiments}
\label{sec: experiments}
% We evaluate and provide comparisons for the model approach compared with other state-of-the art methods. We evaluate the fidelity of the model using both the point forecast accuracy and probabilistic accuracy. It will follow that probabilistic loss allows use to capture the volatility in timeseries better - especially the total load forecasts which have a high variance and relies on occupancy behavior, making it highly stochastic.

We benchmark the performance of some of the prominent models today following our problem formulation. 
% We compare the performance of \modelname to those benchmarks. 
Specifically, in this work, we evaluate and compare the zero-shot performance of Encoder LSTM (LSTM-E) \citep{genTL}, Encoder \& Decoder LSTM (LSTM) \cite{multizoneIndoorTemp},  Transformer \cite{attentionIsAllYouNeed}, TimeXer \cite{timexer}, TSMixer \cite{TSMixer}, and PatchTST \cite{patchTST} models. Some models are further modified to have a robust version, specifically LSTM-R, Transformer-R, and TimeXer-R, and are created to handle unavailable data and provide benchmarks for robust forecasting.  
% The model choices are chosen to reflect the commonly used models for benchmarking in building energy and analytics community as well as in machine learning time series forecasting community, to bridge the understanding of the results for both of them for understanding how these models fare for cross-domain generalization in residential setting. 
For all experiments, our context length - the length of past timesteps used in input - is $L_c=384$, equivalent to 4 days of data, and we predict $L_f=96$ timesteps which equivalent to 1 day of data. This context and horizon is consistent with common short-term load residential forecasting settings~\cite{buildingsBench, re_alden, he_maml}. The masking operator $\mathcal{M}$ used in our experiments drops \textit{whole input channels} and individual home properties with high probabilities (Appendix~\ref{app:masking}). 
% Each input co variate gets a independent high probability of being dropped. All of the home properties share a random variable of dropping the probabilities.  
We evaluate forecasting performance using two complementary metrics: Normalized Root Mean Square Error (NRMSE) and Continuous Ranked Probability Score (CRPS), corresponding to evaluating point prediction accuracy and probabilistic prediction accuracy (Appendix \ref{sec: metrics_used}).

\subsection{In-Domain Evaluation}
Firstly, we evaluate the models in-domain with full data availability and observe their accuracy. Then we develop the top four performing model as robust models. Robust models are then evaluated in-domain under conditions where some of the dynamic and static covariates are missing. Here, we consider the past target is always available.

For in domain evaluation, we train models on a data split consisting of homes from US states from all climate zones, and test it on data from the same homes (Appendix \ref{sec:domain_descriptions}). In-distribution evaluations provide a preliminary understanding of how different models may perform under stronger domain shifts. The results are presented in Table \ref{tab:combined_results}.

\begin{table}[h]
\small
\centering
\caption{Performance on three forecasting tasks on the US States I dataset, comparing models under full data availability and robust models under missing/unavailable data.}
\label{tab:combined_results}
\begin{tabular}{ll|ccc|ccc}
\toprule
 & & \multicolumn{3}{c|}{NRMSE} & \multicolumn{3}{c}{CRPS} \\
Setting & Model & Total & HVAC & Temp & Total & HVAC & Temp \\
\midrule
\multirow{6}{3cm}{\centering Full data availability}
 & LSTM-E       & 46.47 & 46.21 & 2.68 & 0.143 & 0.090 & 0.802 \\
 & LSTM         & \underline{40.39} & {39.48} & 3.97 & \underline{0.128} & 0.087 & 1.407 \\
 & Transformer  & \textbf{38.64} & \underline{34.45} & 3.78 & \textbf{0.121} & \textbf{0.077} & 1.501 \\
 & TSMixer      & 42.08 & 35.94 & \underline{1.19} & 0.166 & 0.091 & 0.364 \\
 & TimeXer      & 40.43 & \textbf{32.56} & \textbf{0.98} & 0.156 & \underline{0.081} & \textbf{0.263} \\
 & PatchTST     & 44.02 & 38.02 & 1.21 & 0.147 & 0.094 & \underline{0.325} \\
\midrule
\multirow{4}{3cm}{\centering Missing and/or unavailable data \\ (\textbf{-R}: robust models)}
 & LSTM-R         & 46.84 & 44.27 & 5.33 & 0.191 & 0.102 & 2.134 \\
 & Transformer-R  & 48.48 & 42.42 & 5.06 & 0.194 & 0.098 & 1.994 \\
 & TSMixer-R      & \underline{43.49} & \underline{37.29} & \underline{1.21} & \underline{0.174} & \underline{0.094} & \underline{0.367} \\
 & TimeXer-R      & \textbf{42.13} & \textbf{35.11} & \textbf{0.95} & \textbf{0.163} & \textbf{0.086} & \textbf{0.251} \\
\bottomrule
\end{tabular}
\end{table}

We observe transformer and TimeXer perform better when all data is available in-domain. Many works in building analytics community have previously focused on such in-domain setting with full data availability. But under conditions with missing data, TimeXer and TSMixer models consistently out-perform classical transformer and LSTMs. As LSTM, Transformer, TSMixer and TimeXer are the best performing model in-domain, we use their robust variants for the following experiments.

\subsection{Cross-Domain Zero-Shot Evaluation}
We evaluate how well models generalize when trained on one group of homes and tested on a different group of homes categorized under a specific domain-shift. To do this, we define four \emph{domain categories}, each capturing a distinct axis of heterogeneity in the residential stock: state-based geography (State), ASHRAE climate zone (CZ), wall construction type (Wall), and heating technology (Heating). Within each category, we partition homes into four to five \emph{domain splits}. For example, the State category contains splits for Midwestern, Northeastern, Southern, Mountain West, and South-Central states. Full details of all categories and splits are provided in Section~\ref{sec:domain_descriptions}.

For each category, we run a leave-one-split-out evaluation using the robust model variants. We train on a single split and measure zero-shot accuracy on each of the remaining splits in that category, then repeat with every other split serving as the training source. Each training split draws 400 randomly sampled homes and each test split draws 100 randomly sampled homes. Table~\ref{tab:forecasting_results} reports the mean accuracy across every source–target pair within each category.

\begin{table}[htbp]
\centering
\caption{Forecasting results with standard deviation across total load, HVAC load, and indoor temperature forecasting tasks. Best values are in \textbf{bold}, second-best are \underline{underlined}.}
\setlength{\tabcolsep}{4pt}
\small
\begin{tabular}{lllcccc}
\toprule
Task & Metric & Model & State & CZ & Wall & Heating \\
\midrule
\multirow{8}{*}{\shortstack{Total\\Load}} 
& \multirow{4}{*}{NRMSE} 
& LSTM-R        & 46.878 $\pm$ 2.981 & 47.926 $\pm$ 3.266 & 48.082 $\pm$ 2.152 & 46.546 $\pm$ 6.374 \\
& & Transformer-R & 49.774 $\pm$ 3.372 & 50.574 $\pm$ 3.439 & 49.713 $\pm$ 2.575 & 48.527 $\pm$ 6.683 \\
& & TSMixer-R     & \textbf{42.892 $\pm$ 2.266} & \textbf{44.363 $\pm$ 2.189} & \textbf{44.959 $\pm$ 2.273} & \textbf{45.249 $\pm$ 2.324} \\
& & TimeXer-R     & \underline{43.586 $\pm$ 5.288} & \underline{47.884 $\pm$ 6.516} & \underline{48.261 $\pm$ 6.209} & \underline{47.554 $\pm$ 6.025} \\
\cmidrule(lr){2-7}
& \multirow{4}{*}{CRPS} 
& LSTM-R        & 0.197 $\pm$ 0.008 & 0.200 $\pm$ 0.006 & 0.180 $\pm$ 0.011 & 0.205 $\pm$ 0.023 \\
& & Transformer-R & 0.203 $\pm$ 0.011 & 0.199 $\pm$ 0.005 & 0.184 $\pm$ 0.012 & 0.204 $\pm$ 0.029 \\
& & TSMixer-R     & \textbf{0.173 $\pm$ 0.006} & \textbf{0.175 $\pm$ 0.005} & \textbf{0.165 $\pm$ 0.009} & \textbf{0.166 $\pm$ 0.008} \\
& & TimeXer-R     & \underline{0.174 $\pm$ 0.022} & \underline{0.186 $\pm$ 0.028} & \underline{0.175 $\pm$ 0.027} & \underline{0.172 $\pm$ 0.025} \\
\midrule
\multirow{8}{*}{\shortstack{HVAC\\Load}} 
& \multirow{4}{*}{NRMSE} 
& LSTM-R        & 44.099 $\pm$ 3.520 & 43.765 $\pm$ 4.692 & 44.201 $\pm$ 2.791 & 42.766 $\pm$ 6.457 \\
& & Transformer-R & 44.780 $\pm$ 4.229 & 46.222 $\pm$ 5.158 & 44.535 $\pm$ 3.671 & 44.229 $\pm$ 6.876 \\
& & TSMixer-R     & \textbf{37.423 $\pm$ 1.994} & \textbf{36.405 $\pm$ 2.307} & \textbf{37.217 $\pm$ 1.738} & \underline{38.855 $\pm$ 2.702} \\
& & TimeXer-R     & \underline{38.814 $\pm$ 7.809} & \underline{42.709 $\pm$ 10.206} & \underline{43.136 $\pm$ 9.625} & \textbf{36.817 $\pm$ 2.572} \\
\cmidrule(lr){2-7}
& \multirow{4}{*}{CRPS} 
& LSTM-R        & 0.110 $\pm$ 0.006 & 0.105 $\pm$ 0.005 & 0.100 $\pm$ 0.011 & 0.111 $\pm$ 0.033 \\
& & Transformer-R & 0.111 $\pm$ 0.009 & 0.109 $\pm$ 0.007 & \underline{0.097 $\pm$ 0.012} & 0.112 $\pm$ 0.032 \\
& & TSMixer-R     & \textbf{0.098 $\pm$ 0.004} & \textbf{0.092 $\pm$ 0.004} & \textbf{0.085 $\pm$ 0.009} & \underline{0.089 $\pm$ 0.010} \\
& & TimeXer-R     & \underline{0.100 $\pm$ 0.019} & \underline{0.106 $\pm$ 0.025} & 0.098 $\pm$ 0.025 & \textbf{0.083 $\pm$ 0.010} \\
\midrule
\multirow{8}{*}{\shortstack{Temp-\\erature}} 
& \multirow{4}{*}{NRMSE} 
& LSTM-R        & 5.991 $\pm$ 0.546 & 6.108 $\pm$ 0.674 & 6.301 $\pm$ 1.122 & 7.009 $\pm$ 0.992 \\
& & Transformer-R & 5.535 $\pm$ 0.401 & 5.317 $\pm$ 0.325 & 5.524 $\pm$ 0.329 & 4.642 $\pm$ 1.108 \\
& & TSMixer-R     & \underline{1.292 $\pm$ 0.036} & \underline{1.348 $\pm$ 0.140} & \underline{1.246 $\pm$ 0.096} & \underline{1.341 $\pm$ 0.086} \\
& & TimeXer-R     & \textbf{1.040 $\pm$ 0.031} & \textbf{1.102 $\pm$ 0.106} & \textbf{1.021 $\pm$ 0.091} & \textbf{1.098 $\pm$ 0.075} \\
\cmidrule(lr){2-7}
& \multirow{4}{*}{CRPS} 
& LSTM-R        & 2.410 $\pm$ 0.230 & 2.446 $\pm$ 0.256 & 2.537 $\pm$ 0.343 & 2.896 $\pm$ 0.433 \\
& & Transformer-R & 2.156 $\pm$ 0.153 & 2.040 $\pm$ 0.115 & 2.142 $\pm$ 0.129 & 1.872 $\pm$ 0.392 \\
& & TSMixer-R     & \underline{0.380 $\pm$ 0.006} & \underline{0.412 $\pm$ 0.046} & \underline{0.379 $\pm$ 0.030} & \underline{0.411 $\pm$ 0.029} \\
& & TimeXer-R     & \textbf{0.263 $\pm$ 0.010} & \textbf{0.291 $\pm$ 0.036} & \textbf{0.271 $\pm$ 0.028} & \textbf{0.296 $\pm$ 0.024} \\
\bottomrule
\end{tabular}
\label{tab:forecasting_results}
\end{table}
For total and HVAC load forecasting, TSMixer-R consistently achieves the lowest NRMSE and CRPS across all domain categories, followed by TimeXer-R. The Wall construction domain yields the lowest CRPS for both tasks, consistent with its relatively low JS divergence values (Table \ref{tab:walls_dist_js}), suggesting envelope-driven shifts are more structured. The Heating domain introduces the highest variance across models, indicating that heating technology shifts produce more heterogeneous cross-domain behavior. HVAC errors are consistently lower than total load errors because HVAC load is more tightly coupled to the weather features, and inherently have much lower variance compared to total loads. As load values are influenced by exogenous features, MLP-mixer model see those features multiple times when mixing, learning better dependence compared to a single attention operation.

Indoor temperature forecasting presents a different picture: TimeXer-R achieves the best accuracy, while LSTM-R and Transformer-R degrade by nearly an order of magnitude. TimeXer performs self attention on target, helping  maintain the thermal inertia, and is still given some influence with token based cross-attention. This balances thermal inertia and exogenous influence better, which is what temperature dynamics require. This suggests cross-attention-based and mixer-style architectures better capture smooth, setpoint-driven thermal dynamics and exogenous driven load change better, respectively, under domain shift, whereas recurrent models and vanilla transformers without architectural conditioning of those exogenous features have high difficulty in generalizing across construction and climate changes. While intuitive, though not trivial or implied, we do also see a correlation in NRMSE and CRPS values - a model having one lower metric also has a lower other metric.

\subsection{Zero-Shot Evaluation on Real data}
\label{subsec: real_data_eval}
We evaluate the robust model variants zero-shot on five real-world residential datasets — ECOBEE, HEAPO, IDEAL, NEST, and REFIT — consisting of all tasks and present the results in Table \ref{tab:real_data_results}. We train on data split US States I (see section \ref{sec:domain_descriptions}) where 400 homes are sampled, to make up the training domain's distribution for this experiment.  

When tested on real data, we first observe that point forecast accuracy (NRMSE) degrades substantially relative to the synthetic domain evaluations, which is expected given the sim-to-real gap and the absence of any target-domain fine-tuning. TSMixer-R achieves the lowest NRMSE in six of eight columns, with TimeXer-R competitive on temperature tasks. Recurrent and vanilla transformer variants lag behind in generalizing to real data. This observation reinforces our observation made in previous experiment regarding Mixer style model aiding in seeing exogenous features more for load forecasting, and TimeXer finding a balance between self-attention and exogenous influence which is what temperature dynamics follow.

It is also observed that probabilistic performance has much greater success compared to point estimates. CRPS values are quite lower and more consistent across datasets, indicating that even without fine-tuning, models capture the general distributional shape of the target signal — including load spikes — quite better. This is quite consequential for downstream control and grid optimization tasks, where a calibrated predictive distribution is often more valuable than a sharp point estimate. TSMixer-R leads on CRPS for most total load and temperature columns, while TimeXer-R matches or surpasses it on HVAC load. Together, these results suggest that probabilistic pretraining on synthetic data transfers more readily than point forecast accuracy, motivating future work on distribution alignment methods to close the remaining sim-to-real gap.

\begin{table*}[htbp]
\small
\centering
\caption{Real-data forecasting results grouped by target. Best values per column within each metric are \textbf{bolded}, second best \underline{underlined}.}
\setlength{\tabcolsep}{4pt}
\begin{tabular}{llcccccccc}
\toprule
& & \multicolumn{4}{c}{Total Load} & HVAC & \multicolumn{3}{c}{Indoor Temperature} \\
\cmidrule(lr){3-6} \cmidrule(lr){7-7} \cmidrule(lr){8-10}
Metric & Model & Heapo & Ideal & Nest & Refit & Nest & Ecobee & Ideal & Nest \\
\midrule
\multirow{4}{*}{NRMSE} 
& LSTM-R        & 89.753 & \underline{97.706} & \underline{95.169} & 100.841 & 82.146 & 5.295 & 7.369 & 1.839 \\
& Transformer-R & 87.902 & 103.560 & 96.067 & 100.473 & 77.994 & 4.328 & 7.029 & 4.980 \\
& TSMixer-R     & \textbf{75.595} & \textbf{92.272} & 95.718 & \textbf{88.860} & \underline{76.232} & \underline{1.859} & \textbf{3.687} & \textbf{1.614} \\
& TimeXer-R     & \underline{81.335} & 116.275 & 205.769 & \underline{99.296} & \textbf{75.933} & \textbf{1.822} & \underline{3.786} & \underline{1.626} \\
\midrule
\multirow{4}{*}{CRPS} 
& LSTM-R        & \underline{0.182} & \textbf{0.049} & \underline{0.023} & \textbf{0.056} & 0.050 & 2.200 & 3.061 & 0.878 \\
& Transformer-R & 0.185 & 0.054 & \textbf{0.022} & \underline{0.057} & 0.050 & 1.771 & 2.476 & 1.889 \\
& TSMixer-R     & \textbf{0.169} & \underline{0.050} & 0.031 & 0.060 & \underline{0.049} & \underline{0.760} & \textbf{1.448} & \textbf{0.719} \\
& TimeXer-R     & 0.204 & 0.096 & 0.095 & 0.091 & \textbf{0.049} & \textbf{0.757} &\underline{1.465} & \underline{0.761} \\
\bottomrule
\end{tabular}
\label{tab:real_data_results}
\end{table*}

\section{Limitations and Future Works}
\label{sec: limitations_and_future}
RESCAST-100K relies on physics-based simulation and still cannot fully replicate real occupant stochasticity and appliance heterogeneity — the primary driver of the sim-to-real gap observed in Section \ref{subsec: real_data_eval}. But the goal, and the future works to address this, includes closer data alignment in the source dataset, which can be achieved through a learned or defined transfer utility function. Having a closer source distribution can aid in having a better prior for sim-to-real transfer. The included real-world datasets are also geographically limited and exhibit significant covariate missingness, motivating broader real-data integration in future releases. While the covariate missingness cannot be directly addressed, we aim to include more real datasets across further geographical locations for better understanding of real domain adaptation. The current benchmarks cover forecasting generalization and integration with downstream control tasks are natural extensions, which highly encourage users to pursue. The next version of \datasetname also plans to include further building envelope details such as U-values, insulation values, etc. that can be used to benchmark physics informed and gray-box forecasting. The dataset also covers only U.S. residential stock;  expanding to international building archetypes would broaden its scope, and require collaboration with different engineering fields.
\section{Conclusion}
\label{sec: conclusion}
We presented RESCAST-100K, a dataset of 15-minute resolution time series for ~100,000 U.S. residential homes, pairing total load, HVAC load, and indoor temperature with exogenous weather signals, setpoints, and over 40 static building covariates. By integrating physics-based simulation with five curated real-world datasets under a unified schema and a configuration-based domain-split interface, RESCAST-100K enables rigorous, reproducible evaluation of transfer learning, domain adaptation, and zero-shot methods for residential forecasting — a class of evaluation no prior resource supported at this scale. Benchmarking six model families, we find that attention- and mixer-based architectures consistently outperform recurrent baselines under distribution shift, and that probabilistic pretraining on synthetic data transfers more reliably to real homes than point estimation. We release RESCAST-100K to support both the machine learning and building systems communities in advancing cross-domain residential forecasting.

\bibliography{references}
%%%%%%%%%%%%%%%%%%%%%%%%%%%%%%%%%%%%%%%%%%%%%%%%%%%%%%%%%%%%

% %  BEGIN APENDIX
% \newpage
\appendix
\section*{Appendix}

\section{Static Home Properties}
\label{appendix:static_covariates}

Each home in \textsc{ResCast-100K} is described by a set of static covariates drawn from ResStock's
building stock characterization. These subset of covariates are selected to reflect attributes that are \emph{accessible without expert intervention}: they correspond to information a homeowner can typically retrieve from thermostat interfaces, public records, or simple visual inspections---without requiring specialized equipment, energy audits, or engineering expertise. The full set is listed in Table~\ref{tab:static_covariates}.

\begin{table}[htbp]
\centering
\caption{Static covariates included per home in \textsc{ResCast-100K}, grouped by category.}
\label{tab:static_covariates}
\small
\setlength{\tabcolsep}{5pt}
\begin{tabular}{l p{8cm}}
\toprule
\textbf{Category} & \textbf{Properties} \\
\midrule
Climate \& Geography    & ASHRAE/IECC Climate Zone, Building America Climate Zone, Climate Zone 2A Split \\
Location                & County Metro Status, PUMA, PUMA Metro Status, City, Latitude, Longitude \\
Building Geometry       & Building Type, Attic Type, Foundation Type, Floor Area, Number of Stories, Garage, \\
                        & Number of Bedrooms, Number of Occupants, Multi-Family/Single Family \\
Envelope                & Wall Type, Wall Exterior Finish, Door Type, Window Type, Window Areas, \\
                        & Interior Shading, Orientation, Neighbors \\
Insulation              & Ceiling, Floor, Foundation Wall, Rim Joist, Roof, Slab, Wall \\
HVAC \& Heating         & Heating Fuel, Heating System Type, Heating Type and Fuel, Cooling System Type, Shared HVAC System \\
Vintage                 & Construction Vintage, Construction Vintage (ACS) \\
\bottomrule
\end{tabular}
\end{table}
The covariates span roughly seven categories encoding the primary dimensions of heterogeneity in the U.S.residential stock. Except latitude and longitude, which are taken as numerical inputs, all other properties are considered categorical and encoded using an ordinal encoder.  Envelope and insulation fields characterize the thermal properties of the building shell, and how they govern indoor temperature dynamics and HVAC load. HVAC and heating fields capture mechanism-level differences in control logic and efficiency. Vintage encodes era-specific construction practices correlated with insulation levels, air sealing, and equipment efficiency. All fields are stored in a single metadata file indexed by \texttt{bldg\_id}. For real datasets , only a subset of covariates is available; the data loader handles missingness by masking unavailable fields, consistent with the robust training objective in Equation. The code includes a complete data dictionary outlining all the values corresponding to the categorical values for each of the variables.

\section{Domains Used: Description and Motivation}
\label{sec:domain_descriptions}

We define domain categories along four axes: geography, climate, envelope construction,
and heating technology. These categories are intended to induce meaningful distribution
shifts while remaining interpretable. State-based domains capture regional variation in
weather, housing stock, building codes, occupant behavior, and utility context. Climate-zone
domains provide a physics-informed partition that directly affects HVAC operation, seasonal
load profiles, and thermal response. Wall construction domains approximate envelope
properties such as thermal mass, insulation practice, and infiltration pathways. Finally,
heating-system domains capture mechanism-level shifts in control logic, efficiency, and
weather sensitivity across heating technologies. Together, these serve as an example of diverse domains categories to thoroughly evaluate a domain adaptation methodology. The domains are listed in Tables \ref{tab:state_domains}, \ref{tab:climate_zone_domains}, \ref{tab:wall_domains}, and \ref{tab:heating_domains}. The splits create at the beginning of our experimentations, to test the non-robust methods, are also detailed in Table \ref{tab: us_domains} 

Each of these four domain categories corresponds to a distinct research community that has, in isolation, identified the corresponding shift as the dominant generalization barrier. \textbf{State- and region-level shifts} are the focus of utility-side and grid-side forecasting work, where the heterogeneity of weather, building codes, tariffs, and stock composition across regions has motivated cross-region transfer studies on smart-meter and aggregated-demand data~\cite{buildingsBench, cai2019two, antoniadis2024hierarchical, dasshijon2024}. \textbf{Climate-zone shifts} are studied predominantly by the building science and HVAC controls communities, who treat the ASHRAE/IECC partition as the unified domain for thermal-response stratification and have repeatedly shown that models trained in one zone degrade in others without adaptation~\cite{hotDataset, peng2025climate, therml_comfort_transfer, usdoe_climate}. \textbf{Construction- and envelope-type shifts} are studied by the thermal-dynamics modeling community, where wall assembly, insulation, and infiltration determine the dominant time constants of indoor temperature response, and where building-to-building TL is known to be highly sensitive to envelope mismatch~\cite{genTL, li2022building}. \textbf{Heating-equipment shifts} are increasingly studied by the electrification and grid-planning communities, motivated by the projected impact of widespread heat-pump adoption on winter peaks~\cite{lee2025forecasting, chesser2021impact} --- where the COP-, control-, and weather-sensitivity profile of the heating technology determines forecast tractability. No prior dataset enables controlled study of all four and more, and hence requiring \datasetname.

\begin{table}[htbp]
\centering
\small
\caption{Domains curated for representative U.S. states from all IECC climate zones for near in-distribution evaluation of the forecasting models.}
\label{tab: us_domains}
\begin{tabular}{cp{10cm}}
\hline
\textbf{Data Split} & \textbf{States} \\
\hline
US States I &
HI, PR, FL, AZ, TX, GA, KY, NC, IL, CO, MN, ME, AK, ND \\
% \hline
% US States II &
% LA, MS, NM, NV, AL, SC, VA, TN, MA, NY, VT, WI, MT, WY \\
\hline
\end{tabular}
\end{table}

\begin{table}[htbp]
\centering
\caption{State-based geographic domains.}
\label{tab:state_domains}
\small
\begin{tabular}{p{0.25\linewidth}p{0.45\linewidth}p{0.22\linewidth}}
\toprule
\textbf{Domain} & \textbf{Description} & \textbf{States} \\
\midrule
\texttt{state\_northeast} &
Northeast/Mid-Atlantic dense corridor with mixed HVAC regimes. &
MA, CT, RI, NY, NJ, PA, MD, DE, DC, VA \\
\texttt{state\_midwest} &
Great Lakes and Upper Midwest; cold winters and heating-dominant dynamics. &
MI, WI, IL, IN, OH, MN, IA \\
\texttt{state\_southeast} &
Humid Southeast; cooling-heavy with latent loads and long summers. &
FL, GA, SC, NC, AL, MS, TN \\
\texttt{state\_south\_central} &
South-Central and Gulf-adjacent states with hot summers and strong cooling peaks. &
TX, OK, AR, LA, NM \\
\texttt{state\_mountain\_west} &
Mountain and Intermountain West; dry air and large diurnal temperature swings. &
CO, UT, ID, MT, WY, NV, AZ \\
\bottomrule
\end{tabular}
\end{table}

\begin{table}[htbp]
\centering
\caption{Climate-zone domains based on IECC/ASHRAE climate-zone groupings.}
\label{tab:climate_zone_domains}
\small
\begin{tabular}{p{0.28\linewidth}p{0.47\linewidth}p{0.17\linewidth}}
\toprule
\textbf{Domain} & \textbf{Description} & \textbf{Zones} \\
\midrule
\texttt{cz\_hot\_humid} &
Hot-humid climates; cooling-dominant operation with humidity effects. &
1A, 2A \\
\texttt{cz\_hot\_or\_mixed\_dry} &
Hot-dry and mixed-dry climates; sensible loads and large diurnal swings. &
2B, 3B, 4B \\
\texttt{cz\_marine} &
Marine climates; mild temperatures, narrow ranges, and distinct shoulder seasons. &
3C, 4C \\
\texttt{cz\_mixed\_humid} &
Mixed-humid climates; balanced heating and cooling, useful as common transfer sources. &
3A, 4A \\
\texttt{cz\_cold\_to\_very\_cold} &
Cold to very cold climates; heating-dominant regimes and winter dynamics. &
5A, 5B, 6A, 6B, 7A \\
\bottomrule
\end{tabular}
\end{table}

\begin{table}[htbp]
\centering
\caption{Envelope-construction domains defined by wall type and exterior finish.}
\label{tab:wall_domains}
\small
\begin{tabular}{p{0.20\linewidth}p{0.35\linewidth}p{0.10\linewidth}p{0.25\linewidth}}
\toprule
\textbf{Domain} & \textbf{Description} & \textbf{Wall type} & \textbf{Exterior finish} \\
\midrule
\texttt{wood\_siding} &
Lightweight wood-framed envelope with faster thermal response. &
Wood Frame &
Vinyl Light; Wood Medium/Dark; Fiber-Cement Light \\
\texttt{brick\_veneer} &
Wood frame with brick veneer; moderate thermal mass and common archetype. &
Wood Frame &
Brick Medium/Dark; Brick Light \\
\texttt{masonry\_block} &
Masonry or concrete block walls with higher thermal mass and lag. &
Concrete &
Stucco Light; Stucco Medium/Dark \\
\texttt{stucco} &
Stucco exteriors representing a regional envelope archetype shift. &
Wood Frame &
Stucco Light; Stucco Medium/Dark \\
\texttt{structural\_brick} &
Structural brick walls with full masonry, high thermal mass, and strong lag. &
Brick &
None; Stucco Light; Vinyl Light \\
\bottomrule
\end{tabular}
\end{table}

\begin{table}[htbp]
\centering
\caption{Heating-technology domains.}
\label{tab:heating_domains}
\small
\begin{tabular}{p{0.30\linewidth}p{0.44\linewidth}p{0.20\linewidth}}
\toprule
\textbf{Domain} & \textbf{Description} & \textbf{Heating type and fuel} \\
\midrule
\texttt{heat\_electric\_furnace} &
Electric furnace with forced-air distribution. &
Electricity Electric Furnace \\
\texttt{heat\_hp\_electric} &
Electric air-source and mini-split heat-pump homes with variable COP behavior. &
Electricity ASHP; Electricity MSHP \\
\texttt{heat\_elec.\_resistance\_zonal} &
Zonal electric resistance heating with high weather sensitivity. &
Electricity Baseboard; Electricity Electric Wall Furnace \\
\texttt{heat\_electric\_boiler} &
Electric boiler with hydronic distribution. &
Electricity Electric Boiler \\
\bottomrule
\end{tabular}
\end{table}

\begin{figure}[htbp]
    \centering
    \fbox{\includegraphics[width=\textwidth]{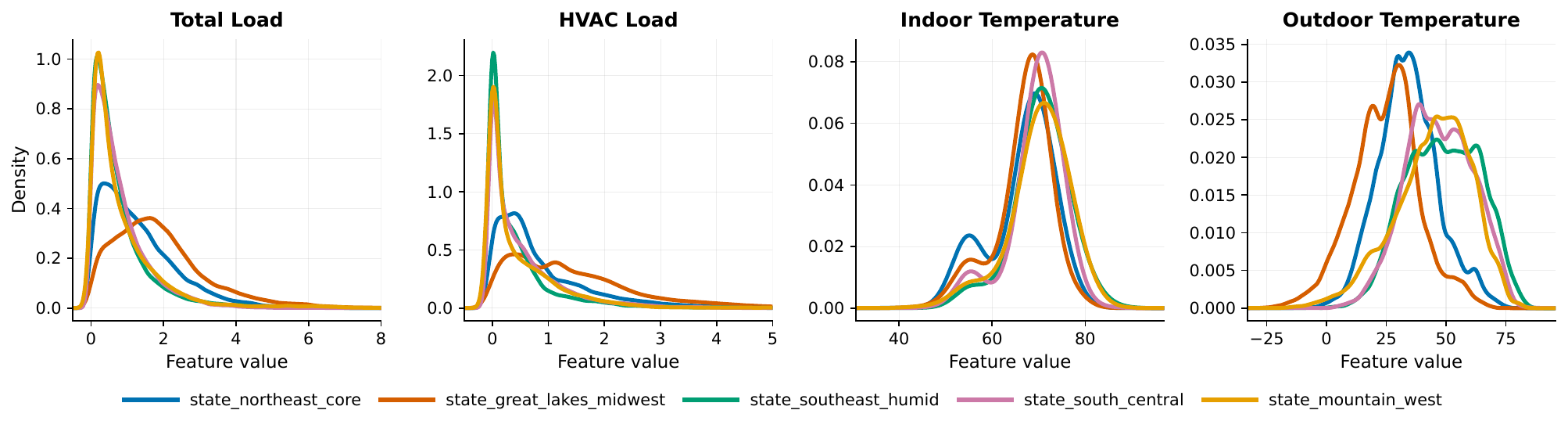}}
    \caption{Distribution of total load, HVAC load, indoor and outdoor temperatures across state domains.}
    \label{fig:state_dist_viz}
\end{figure}

\begin{figure}[htbp]
    \centering
    \fbox{\includegraphics[width=\textwidth]{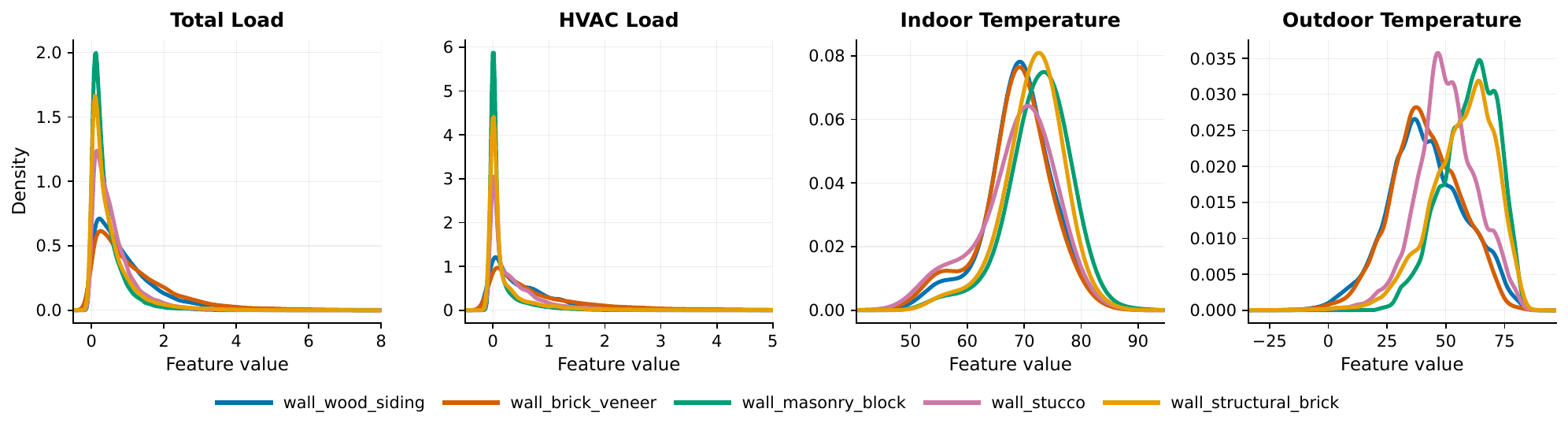}}
    \caption{Distribution of total load, HVAC load, indoor and outdoor temperatures across different wall type domains.}
    \label{fig:walls_dist_viz}
\end{figure}

\begin{figure}[htbp]
    \centering
    \fbox{\includegraphics[width=\textwidth]{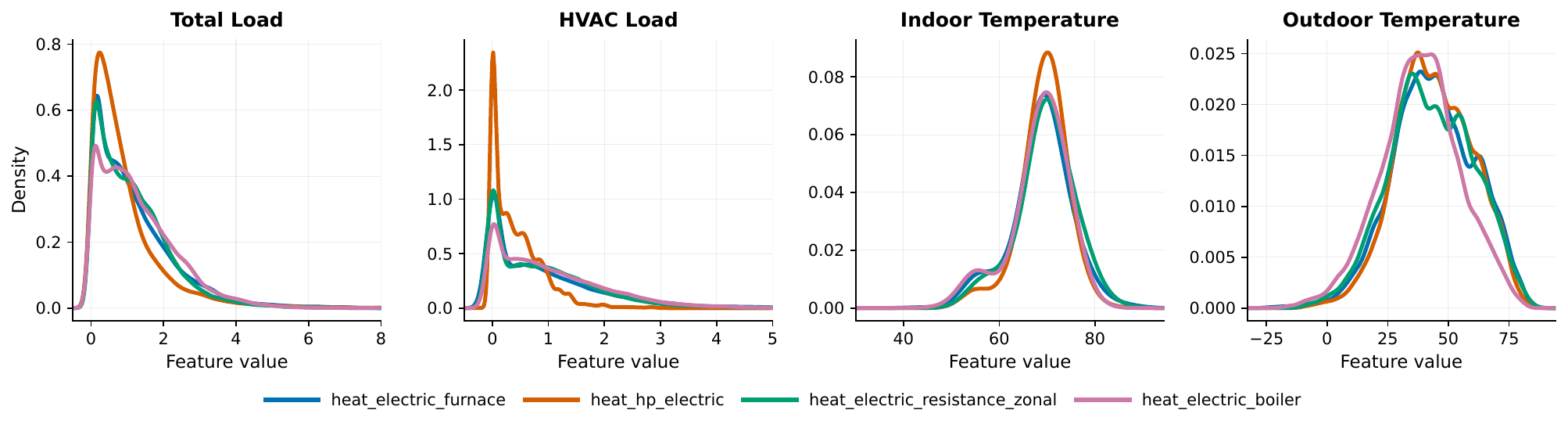}}
    \caption{Distribution of total load, HVAC load, indoor and outdoor temperatures across heating equipment domains.}
    \label{fig:heat_dist_viz}
\end{figure}

We compare the domain shifts across domains under each category. We compare the target values and outdoor temperature. While these are a great way to visualize the data distributions, they do not capture the complexity of time series due to factors such as dropping the time indices leading, failing to capture the rate of changes, the conditional distributions, and more. Nonetheless, they are a gateway to understanding those underlying distributions. The distributions are visualized in Figures \ref{fig:state_dist_viz}, 
% \ref{fig:cz_dist_viz}, 
\ref{fig:walls_dist_viz}, and \ref{fig:heat_dist_viz}. We also quantify these use JS divergence in tables \ref{tab:state_dist_js}, \ref{tab:cz_dist_js}, \ref{tab:walls_dist_js}, and \ref{tab:heat_dist_js}.

\begin{figure}[htbp]
    \centering
    \fbox{\includegraphics[width=\textwidth]{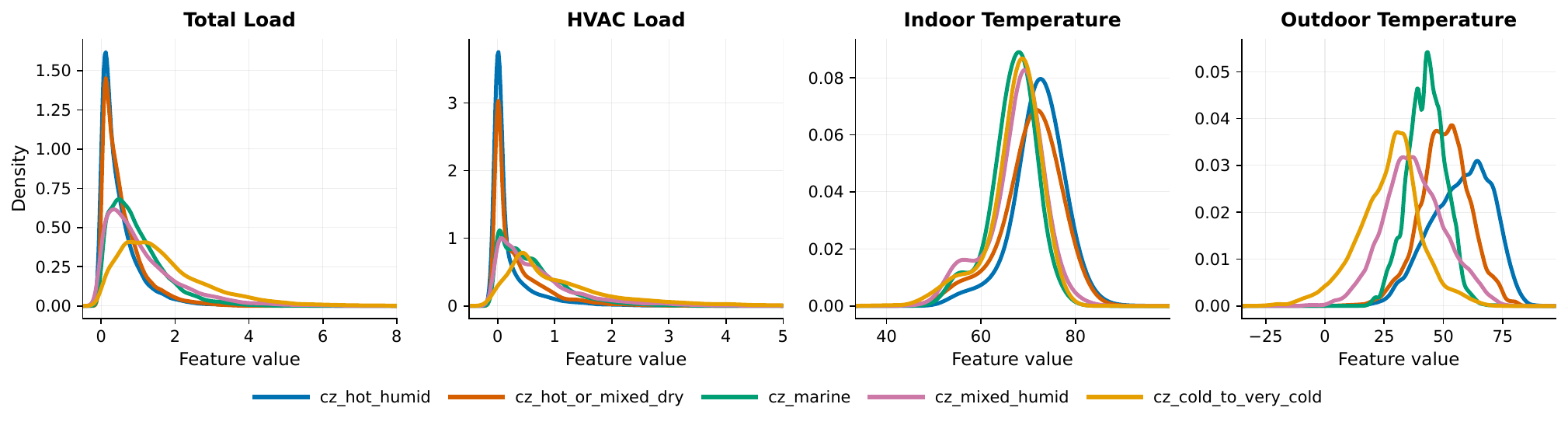}}
    \caption{Distribution of total load, HVAC load, indoor and outdoor temperatures across climate zone domains.}
    \label{fig:cz_dist_viz}
\end{figure}

\begin{table}[htbp]
\centering
\small
\caption{JS Divergence between state domains.}
\label{tab:state_dist_js}
\begin{tabular}{llcccc}
\toprule
\textbf{Source Domain} & \textbf{Target Domain} & \textbf{Total Load} & \textbf{HVAC Load} & \textbf{Indoor Temp} & \textbf{Outdoor Temp} \\
\midrule
\texttt{northeast\_core}      & \texttt{g.\_lakes\_midwest} & 0.0363 & 0.0504 & 0.0489 & 0.0690 \\
\texttt{northeast\_core}      & \texttt{southeast\_humid}      & 0.0518 & 0.0790 & 0.1145 & 0.1285 \\
\texttt{northeast\_core}      & \texttt{south\_central}        & 0.0410 & 0.0457 & 0.0748 & 0.1119 \\
\texttt{northeast\_core}      & \texttt{mountain\_west}        & 0.0484 & 0.0756 & 0.1366 & 0.0965 \\
\texttt{g.\_lakes\_midwest}& \texttt{southeast\_humid}      & 0.1542 & 0.1792 & 0.1400 & 0.2585 \\
\texttt{g.\_lakes\_midwest}& \texttt{south\_central}        & 0.1403 & 0.1331 & 0.1150 & 0.2480 \\
\texttt{g.\_lakes\_midwest}& \texttt{mountain\_west}        & 0.1350 & 0.1559 & 0.1524 & 0.1950 \\
\texttt{southeast\_humid}     & \texttt{south\_central}        & 0.0031 & 0.0110 & 0.0348 & 0.0100 \\
\texttt{southeast\_humid}     & \texttt{mountain\_west}        & 0.0068 & 0.0103 & 0.0212 & 0.0292 \\
\texttt{south\_central}       & \texttt{mountain\_west}        & 0.0101 & 0.0103 & 0.0447 & 0.0186 \\
\bottomrule
\end{tabular}
\end{table}

\begin{table}[htbp]
\centering
\small
\caption{JS Divergence between climate zone domains.}
\label{tab:cz_dist_js}
\begin{tabular}{llcccc}
\toprule
\textbf{Source Domain} & \textbf{Target Domain} & \textbf{Total Load} & \textbf{HVAC Load} & \textbf{Indoor Temp} & \textbf{Outdoor Temp} \\
\midrule
\texttt{hot\_humid}           & \texttt{hot\_or\_mixed\_dry}   & 0.0074 & 0.0205 & 0.0394 & 0.0650 \\
\texttt{hot\_humid}           & \texttt{marine}                & 0.0818 & 0.1309 & 0.2553 & 0.2330 \\
\texttt{hot\_humid}           & \texttt{mixed\_humid}          & 0.0762 & 0.1283 & 0.2067 & 0.2286 \\
\texttt{hot\_humid}           & \texttt{cold\_to\_very\_cold}  & 0.2074 & 0.2440 & 0.2726 & 0.4134 \\
\texttt{hot\_or\_mixed\_dry}  & \texttt{marine}                & 0.0533 & 0.0579 & 0.1641 & 0.1038 \\
\texttt{hot\_or\_mixed\_dry}  & \texttt{mixed\_humid}          & 0.0597 & 0.0607 & 0.1146 & 0.1387 \\
\texttt{hot\_or\_mixed\_dry}  & \texttt{cold\_to\_very\_cold}  & 0.1810 & 0.1630 & 0.1694 & 0.3411 \\
\texttt{marine}               & \texttt{mixed\_humid}          & 0.0145 & 0.0092 & 0.0452 & 0.0931 \\
\texttt{marine}               & \texttt{cold\_to\_very\_cold}  & 0.0634 & 0.0584 & 0.0516 & 0.2466 \\
\texttt{mixed\_humid}         & \texttt{cold\_to\_very\_cold}  & 0.0460 & 0.0463 & 0.0403 & 0.0837 \\
\bottomrule
\end{tabular}
\end{table}

\begin{table}[htbp]
\centering
\small
\caption{JS Divergence between wall construction domains.}
\label{tab:walls_dist_js}
\begin{tabular}{llcccc}
\toprule
\textbf{Source Domain} & \textbf{Target Domain} & \textbf{Total Load} & \textbf{HVAC Load} & \textbf{Indoor Temp} & \textbf{Outdoor Temp} \\
\midrule
\texttt{wood\_siding}        & \texttt{brick\_veneer}        & 0.0064 & 0.0112 & 0.0291 & 0.0077 \\
\texttt{wood\_siding}        & \texttt{masonry\_block}       & 0.0877 & 0.1189 & 0.1295 & 0.1830 \\
\texttt{wood\_siding}        & \texttt{stucco}               & 0.0357 & 0.0332 & 0.0470 & 0.0615 \\
\texttt{wood\_siding}        & \texttt{structural\_brick}    & 0.0555 & 0.0885 & 0.1041 & 0.1208 \\
\texttt{brick\_veneer}       & \texttt{masonry\_block}       & 0.1185 & 0.1601 & 0.1660 & 0.2022 \\
\texttt{brick\_veneer}       & \texttt{stucco}               & 0.0606 & 0.0608 & 0.0581 & 0.0633 \\
\texttt{brick\_veneer}       & \texttt{structural\_brick}    & 0.0800 & 0.1253 & 0.1315 & 0.1385 \\
\texttt{masonry\_block}      & \texttt{stucco}               & 0.0218 & 0.0427 & 0.0841 & 0.0798 \\
\texttt{masonry\_block}      & \texttt{structural\_brick}    & 0.0082 & 0.0058 & 0.0267 & 0.0131 \\
\texttt{stucco}              & \texttt{structural\_brick}    & 0.0119 & 0.0281 & 0.0603 & 0.0448 \\
\bottomrule
\end{tabular}
\end{table}

\begin{table}[htbp]
\centering
\small
\caption{JS Divergence between heating system domains.}
\label{tab:heat_dist_js}
\begin{tabular}{llcccc}
\toprule
\textbf{Source Domain} & \textbf{Target Domain} & \textbf{Total Load} & \textbf{HVAC Load} & \textbf{Indoor Temp} & \textbf{Outdoor Temp} \\
\midrule
\texttt{electric\_furnace}              & \texttt{hp\_electric}                 & 0.0168 & 0.0794 & 0.0247 & 0.0050 \\
\texttt{electric\_furnace}              & \texttt{elec.\_res.\_zonal}  & 0.0039 & 0.0038 & 0.0213 & 0.0065 \\
\texttt{electric\_furnace}              & \texttt{electric\_boiler}             & 0.0087 & 0.0105 & 0.0479 & 0.0170 \\
\texttt{hp\_electric}                   & \texttt{elec.\_res.\_zonal}  & 0.0188 & 0.0848 & 0.0408 & 0.0072 \\
\texttt{hp\_electric}                   & \texttt{electric\_boiler}             & 0.0293 & 0.0980 & 0.0394 & 0.0207 \\
\texttt{elec.\_res.\_zonal}    & \texttt{electric\_boiler}             & 0.0065 & 0.0081 & 0.0555 & 0.0149 \\
\bottomrule
\end{tabular}
\end{table}

\section{Loss function, Evaluation Metric, and Masking Details}
\label{app:problem-details}

This appendix specifies some of the components of Section~\ref{sec: problem_statement} and Section~\ref{sec: experiments}: the training loss $\mathcal{L}_{\mathrm{prob}}$, the evaluation metrics, and the masking distribution $\mathcal{P}_M$.

\subsection{Probabilistic Loss}
\label{app:loss}

Let $\rho_\tau(\hat{q}, y) = (\tau - \mathbf{1}[y < \hat{q}])(\hat{q} - y)$ denote the pinball (quantile) loss at level $\tau \in (0,1)$. Given a predicted quantile tensor $\hat{\mathbf{Q}}_\theta \in \mathbb{R}^{L_f \times Q}$ from Eq.~\eqref{eq:predictor} and ground truth $\mathbf{y}_{1:L_f}$, we define the training loss as the average pinball loss over horizon steps and quantile levels:
\begin{equation}
    \mathcal{L}_{\mathrm{prob}}(\hat{\mathbf{q}}_\theta, \mathbf{y}_{1:L_f}) \;=\; \frac{1}{L_f \cdot Q} \sum_{t=1}^{L_f} \sum_{q=1}^{Q} \rho_{\tau_q}\bigl(\hat{Q}_{t,\tau_q},\, y_t\bigr).
    \label{eq:pinball-app}
\end{equation}
We use $Q = 10$ quantile levels uniformly spaced on $[0.05, 0.95]$, i.e.\ $\tau_q \in \{0.05, 0.15, \dots, 0.95\}$.

\subsection{Evaluation Metrics}
\label{sec: metrics_used}
NRMSE measures point forecast accuracy by computing the root mean squared error normalized by the mean of the target variable, making it comparable across tasks with different scales:
\begin{equation}
    \text{NRMSE} = \frac{\sqrt{\frac{1}{L_f}\sum_{t=1}^{L_f}(\hat{y}_t - y_t)^2}}{\bar{y}} 
    \times 100,
\end{equation}
where $\hat{y}_t$ is the predicted value, $y_t$ is the ground truth, and $\bar{y}$ is the mean of the target series. CRPS evaluates probabilistic forecast quality by measuring the compatibility of a predicted distribution with an observed outcome, rewarding both calibration and sharpness. Since our model outputs a finite set of quantile predictions, we approximate CRPS via the sum of pinball losses across quantile levels:
\begin{equation}
    \text{CRPS} \approx \frac{2}{L_f \cdot Q} \sum_{t=1}^{L_f} \sum_{q=1}^{Q} 
    \rho_{\tau_q}(\hat{Q}_{t,\tau_q}, y_t),
\end{equation}
where $\tau_1, \ldots, \tau_Q$ are the quantile levels and $\rho_{\tau}(z, y) = (\tau - \mathds{1}[y < z])(z - y)$ is the pinball loss at  quantile level $\tau$. In all our experiments we use 10 quantiles uniformly spread between [0.05, 0.95].

\subsection{Masking Operator}
\label{app:masking}

To evaluate robustness to incomplete sensing, we introduce a structured masking operator that randomly removes subsets of input covariates before they are passed to the model. The goal is to simulate realistic missing-data conditions: in residential energy systems, weather feeds may be unavailable, thermostat or HVAC setpoint measurements may be missing, and static home attributes may be partially observed or unavailable. Rather than masking individual entries independently, our operator masks semantically meaningful groups of variables, making it closer to real conditions.

We partition the dynamic-input channels into weather and setpoint sub-blocks. Writing $\mathbf{x}_{1:L_c} = [\mathbf{y}_{1:L_c};\, \mathbf{w}_{1:L_c};\, \mathbf{h}_{1:L_c}]$ and $\mathbf{u}_{1:L_f} = [\mathbf{w}_{1:L_f};\, \mathbf{h}_{1:L_f}]$, where $\mathbf{w}$ stacks the $N_w$ weather channels and $\mathbf{h}$ stacks the $N_h$ HVAC setpoint channels, the masking operator
\[
\mathcal{M}:(\mathbf{x}_{1:L_c},\, \mathbf{u}_{1:L_f},\, \mathbf{s}) \;\mapsto\; (\tilde{\mathbf{x}}_{1:L_c},\, \tilde{\mathbf{u}}_{1:L_f},\, \tilde{\mathbf{s}})
\]
leaves the past target $\mathbf{y}_{1:L_c}$ untouched and acts on the remaining components as follows.

\paragraph{Weather channels.} For each weather variable $j \in \{1,\dots,N_w\}$, we draw a single Bernoulli mask
$z^{w}_{j} \sim \mathrm{Bernoulli}(p_w)$
and apply it jointly to the past and future windows of that channel:
\[
\tilde{\mathbf{w}}_{1:L_c,\,j} \;=\; (1-z^w_j)\,\mathbf{w}_{1:L_c,\,j}, \qquad
\tilde{\mathbf{w}}_{1:L_f,\,j} \;=\; (1-z^w_j)\,\mathbf{w}_{1:L_f,\,j}.
\]
A channel is therefore either fully observed or fully unobserved across the entire $L_c + L_f$ window.

\paragraph{HVAC setpoints.} Heating and cooling setpoints are masked in tandem. A single Bernoulli draw
$z^{h} \sim \mathrm{Bernoulli}(p_h)$
removes all $N_h$ setpoint channels jointly across both the past and future windows when $z^h = 1$:
\[
\tilde{\mathbf{h}}_{1:L_c} = (1-z^h)\,\mathbf{h}_{1:L_c}, \qquad
\tilde{\mathbf{h}}_{1:L_f} = (1-z^h)\,\mathbf{h}_{1:L_f}.
\]

\paragraph{Static house properties.} We partition the static-feature indices $\{1,\dots,N_s\}$ into a set of key attributes $K$ and remaining attributes $R$. Key attributes — which encode information typically retrievable from a thermostat or property record — are dropped with low probability, while remaining attributes are dropped with higher probability:
\[
z^{s}_{i} \sim
\begin{cases}
\mathrm{Bernoulli}(p_{\mathrm{key}}), & i \in K,\\[2pt]
\mathrm{Bernoulli}(p_{\mathrm{other}}), & i \in R,
\end{cases}
\qquad
\tilde{s}_i = (1-z^s_i)\,s_i.
\]

\paragraph{Observation indicator.} Alongside $(\tilde{\mathbf{x}}, \tilde{\mathbf{u}}, \tilde{\mathbf{s}})$, the model receives a binary observation indicator $\mathbf{m}$ with $m = 1$ on observed entries and $m = 0$ on masked entries. This allows the model to distinguish genuinely missing inputs from valid observations rather than conflating either case with a numeric zero. Together, $\mathcal{M}$ and $\mathbf{m}$ form a operator for realistic deployment settings where sensors, forecasts, or building metadata are usually only partially available.

\section{Forecasting Model Suite}
\label{app:model_suite}

\subsection{Model Families and Motivation}
\label{app:model_families}

We evaluate a set of recurrent, attention-based, and modern long-horizon time-series models in order to connect model families commonly used in building analytics with those currently used in the machine learning forecasting community. Let $L=384$ denote the historical context length and $H=96$ denote the forecast horizon. At each prediction time, the model receives historical target values $y_{1:L}$, historical weather covariates $w_{1:L}$, historical HVAC setpoint covariates $u_{1:L}$, known future weather covariates $w_{L+1:L+H}$, known future HVAC setpoint covariates $u_{L+1:L+H}$, and static building attributes $s$ when used by the model.

\paragraph{Encoder LSTM (LSTM-E).}
LSTM-E corresponds to the baseline encoder-only LSTM. It follows the classical many-to-one recurrent forecasting design: an LSTM encoder reads the past window and the final hidden state is mapped directly to the full forecast horizon. This architecture is included because encoder LSTMs remain a common baseline in building load and indoor temperature forecasting, and provide a simple recurrent reference point against which more structured encoder--decoder and attention models can be compared \citep{genTL}.

\paragraph{Encoder--Decoder LSTM (LSTM).}
LSTM uses an encoder--decoder LSTM structure: a bidirectional LSTM encoder summarizes the historical sequence and a decoder LSTM consumes known future covariates to produce one prediction per future time step. Static building properties are used to condition the encoded state. This model represents the stronger recurrent family often used in building analytics, where known future weather and control signals are naturally incorporated into the decoder \citep{multizoneIndoorTemp}.

\paragraph{Transformer.}
Transformer is an encoder--decoder Transformer in which past target, weather, and control covariates form the encoder sequence, while known future weather and control covariates form the decoder sequence. We include this model because self-attention is the standard reference architecture for sequence modeling in modern ML \citep{attentionIsAllYouNeed}, and because its ability to model long-range temporal dependencies is relevant for building thermal and energy dynamics.

\paragraph{TimeXer.}
TimeXer is a recent forecasting architecture designed to combine endogenous temporal patches with exogenous covariate information \citep{timexer} through modified cross attention. It is included to represent recent ML forecasting models that explicitly target prediction with external variables.

\paragraph{TSMixer.}
TSMixer uses MLP-style mixing across temporal and feature dimensions rather than recurrent or attention-based sequence processing \citep{TSMixer}. We include it as a lightweight modern ML baseline, since mixer-style models often provide strong performance with lower architectural complexity.

\paragraph{PatchTST.}
PatchTST tokenizes time series into patches and applies Transformer-style modeling over patch tokens \citep{patchTST}. It is included because patch-based Transformers are a strong and widely used modern baseline for long-horizon time-series forecasting.

Together, these models bridge two communities: recurrent encoder and encoder--decoder models that are familiar in building analytics, and Transformer, patch-based, and mixer-based architectures that are now common in the broader ML time-series literature.

\subsection{Robust Model Construction}
\label{app:robust_models}

We construct robust versions of selected models to handle unavailable covariates: LSTM-R, Transformer-R, TimeXer-R, and TSMixer-R. These models are chosen as they perform the best under in domain evaluation. Robustness is introduced through structured missingness augmentation and model-specific missing-value handling.

\paragraph{Structured missingness augmentation.}
During robust training, selected inputs are replaced with \texttt{NaN}. Each weather channel is masked across both historical and future windows with probability $p_w=0.6$. The HVAC setpoint pair is masked jointly with probability $p_u=0.9$. Static building properties are divided into key and non-key attributes: key attributes are retained with probability $0.9$, while non-key attributes are dropped with probability $0.9$. For unmasked models, this corruption operator is disabled.

For an input block $X \in \mathbb{R}^{T \times C}$, define the observation mask
\begin{equation}
    M_{t,c} = \mathbf{1}\{X_{t,c} \text{ is finite}\},
\end{equation}
and the zero-filled value
\begin{equation}
    \widetilde{X}_{t,c} = M_{t,c}X_{t,c}.
\end{equation}

\paragraph{NaN-aware tokenization for LSTM-R and Transformer-R.}
For LSTM-R and Transformer-R, each scalar variable is encoded together with its observation indicator:
\begin{equation}
    e_{t,c} = \phi\left([\widetilde{X}_{t,c}, M_{t,c}]\right) \in \mathbb{R}^{d},
\end{equation}
where $\phi$ is a two-layer MLP. The variable embeddings are then pooled with the observation mask:
\begin{equation}
    z_t =
    \mathrm{LN}\left(
    \frac{\sum_{c=1}^{C} M_{t,c} e_{t,c}}
    {\max\left(\sum_{c=1}^{C} M_{t,c},1\right)}
    \right).
\end{equation}
Thus unavailable variables receive zero weight, while the model is explicitly informed which values were observed.

\paragraph{LSTM-R.}
LSTM-R applies the NaN-aware tokenizer to both historical and future inputs,
\begin{equation}
    X^p_t = [y_t, w_t, u_t], \qquad
    X^f_h = [w_{L_c+h}, u_{L_c+h}],
\end{equation}
for $h=1,\ldots,L_f$. A bidirectional LSTM encoder processes $z^p_{1:L_c}$, and a decoder LSTM processes $z^f_{1:L_f}$:
\begin{align}
    h_{L_c}, c_{L_c} &= \mathrm{BiLSTM}_{enc}(z^p_{1:L_c}), \\
    d_{1:L_f} &= \mathrm{LSTM}_{dec}(z^f_{1:L_f}; h_{L_c}, c_{L_c}), \\
    \widehat{y}_{h,q} &= W_q d_h + b_q.
\end{align}

\paragraph{Transformer-R.}
Transformer-R uses the same NaN-aware tokenization but replaces the recurrent encoder and decoder with Transformer blocks:
\begin{align}
    m_{1:L_c} &= \mathrm{TransformerEncoder}(z^p_{1:L_c}), \\
    r_{1:L_f} &= \mathrm{TransformerDecoder}(z^f_{1:L_f}, m_{1:L_c}), \\
    \widehat{y}_{h,q} &= W_q r_h + b_q.
\end{align}
No causal decoder mask is used because decoder inputs are known future covariates rather than future target values.

\paragraph{TimeXer-R and TSMixer-R.}
TimeXer-R and TSMixer-R use the same structured masking augmentation, but preserve the original backbone input interface by replacing non-finite values with a sentinel value:
\begin{equation}
    X^{fill}_{t,c} =
    \begin{cases}
    X_{t,c}, & M_{t,c}=1, \\
    c, & M_{t,c}=0.
    \end{cases}
\end{equation}
This informs the backbone about unavailable covariates during training without changing the TimeXer or TSMixer architecture. 
TimeXer-R uses the same TimeXer backbone family but changes the robust configuration in two ways: it is trained with structured missingness, and it uses a shorter patch length ($p=4$ instead of $p=12$). The shorter patch length gives the model finer temporal resolution when forming endogenous tokens, which can help when missing covariates remove some of the exogenous context. TSMixer-R, in contrast, keeps the same mixer-style backbone and patch length as TSMixer. Its robustness comes only from the training distribution: during training, weather, HVAC, and static building covariates are randomly made unavailable, and missing entries are replaced by the model's robust fill pathway before being passed to the mixer. Thus, TimeXer-R modifies both the training corruption process and the temporal tokenization granularity, whereas TSMixer-R isolates the effect of structured missingness training on an otherwise unchanged TSMixer backbone.

\section{Hyperparameters}
\label{app:model_hyperparameters}
Table \label{tab:model_hparams} outlines the model parameters for each model used in this work. For LSTM-E, the encoder-layer count refers to the stacked LSTM encoder and there is no decoder. For LSTM and LSTM-R, encoder and decoder layers refer to LSTM stacks; the ``FFN/token dim'' column denotes the hidden dimension used by the robust-capable variable tokenizer. For Transformer and Transformer-R, heads, encoder layers, decoder layers, and FFN dimension refer to standard Transformer blocks. TimeXer and PatchTST use attention over patched sequences, so patch length is applicable. TSMixer and TSMixer-R do not use attention heads or decoder layers, so these entries are marked with dashes.

\begin{table}[h]
\centering
\caption{Hyperparameters for the forecasting models. A dash indicates that the quantity is not applicable to that architecture.}
\label{tab:model_hparams}
\resizebox{\textwidth}{!}{
\begin{tabular}{lrrrrrrrrrrl}
\toprule
Model & $d_{\mathrm{model}}$ & Heads & Enc. layers & Dec. layers & FFN/token dim & Dropout & $L_c$ & $L_f$ & Patch & Batch & Robust masking \\
\midrule
LSTM-E & 125 & -- & 1 & -- & -- & 0.1 & 384 & 96 & -- & 128 & No \\
LSTM & 256 & -- & 1 & 1 & 256 & 0.1 & 384 & 96 & -- & 128 & No \\
Transformer & 128 & 8 & 1 & 1 & 256 & 0.1 & 384 & 96 & -- & 128 & No \\
TimeXer & 128 & 8 & 1 & -- & 256 & 0.1 & 384 & 96 & 12 & 256 & No \\
TSMixer & 128 & -- & 1 & -- & 256 & 0.1 & 384 & 96 & 12 & 128 & No \\
PatchTST & 128 & 8 & 1 & -- & 256 & 0.1 & 384 & 96 & 12 & 128 & No \\
\midrule
LSTM-R & 128 & -- & 1 & 1 & 256 & 0.1 & 384 & 96 & -- & 256 & Yes \\
Transformer-R & 128 & 8 & 1 & 1 & 256 & 0.1 & 384 & 96 & -- & 256 & Yes \\
TimeXer-R & 128 & 8 & 1 & -- & 512 & 0.1 & 384 & 96 & 4 & 128 & Yes \\
TSMixer-R & 128 & -- & 1 & -- & 256 & 0.1 & 384 & 96 & 12 & 128 & Yes \\
\bottomrule
\end{tabular}
}
\end{table}

\paragraph{Training hyperparameters.}
All neural models are trained with AdamW using learning rate $5\times 10^{-4}$ and weight decay $10^{-2}$. We use cosine annealing over the full training run with minimum learning rate $10^{-4}$, where the schedule length is the number of epochs multiplied by the number of training batches. Gradients are clipped to norm $1.0$. Mixed-precision training is enabled on CUDA when supported. Unless otherwise specified, models are trained for $3$ epochs, with batch sizes given in Table~\ref{tab:model_hparams}. The checkpoint with the best validation performance is retained for evaluation.

\section{Compute Information}
\label{sec: compute_information}
Dataset generation and model training were performed on high performance compute environments. The EnergyPlus simulations for the 100k homes were executed on a high-performance computing cluster using 48x6 CPU cores across 6 nodes. Total simulation compute was approximately 100 CPU-hours. Model training and evaluation were performed on 4 × V100 GPUs with 32 GB system memory. A single training run for the largest model (TimeXer-R) takes approximately 8 minutes per 3 epochs at batch size 128. Each zero-shot cross-domain evaluation (Section 5.2) requires approximately 8 GPU-minutes per source–target domain pair. The full set of experiments reported in Tables 4–6, including all robust and non-robust variants across the four domain categories and sim-to-real evaluations, required approximately 24 total GPU-hours. Preliminary experiments, failed experimetns, and exploration not reported in the paper consumed an additional 300 CPU and 100 GPU hours.

\section{Synthetic to Real Motivation}
\label{sec: real_vs_synth}
Here we compare the distribution of real v/s synthetic datasets. We do this for indoor temperature values and total load values. We compare the distributions of values itself$y_t$, and the the first differences ($\Delta y_t$). For each dataset, we create a dataset with manual intervention. Considering geographical location and home size - we make a data split of synthetic homes. We compare the distribution of the real dataset and the curated synthetic dataset. We observe that the synthetic distributions are very similar to real datasets, hence providing stronger motivation for synthetic source domain alignment for better sim to real transfer.

\begin{table}[htbp]
\centering
\caption{JS divergence across read datasets and corresponding synthetic splits.}
\label{tab:js_divergence}
\begin{tabular}{lllll}
\toprule
\textbf{Dataset} & \textbf{Location} & \textbf{Synthetic Split} & \textbf{Task} & \textbf{JS} \\
\midrule
ECOBEE & Across USA & All States & Temperature forecasting & 0.1108 \\
HEAPO & Canton of Zurich, Switzerland & MN WI MI & Total load forecasting & 0.0491 \\
IDEAL & United Kingdom & WA OR & Temperature forecasting & 0.0630 \\
IDEAL & United Kingdom & WA OR & Total load forecasting & 0.0272 \\
NEST & California, USA & CA & Temperature forecasting & 0.1349 \\
NEST & California, USA & CA & HVAC load forecasting & 0.0871 \\
NEST & California, USA & CA & Total load forecasting & 0.1492 \\
REFIT & United Kingdom & WA OR & Total load forecasting & 0.0188 \\
\bottomrule
\end{tabular}
\end{table}

\begin{figure}[htbp]
\centering

% -------- First row --------
\begin{subfigure}{0.31\linewidth}
\centering
\fbox{\includegraphics[width=\linewidth]{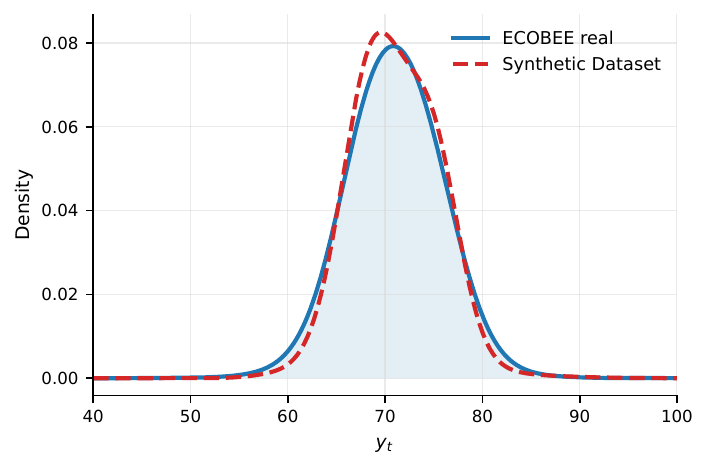}}
\caption{ECOBEE ($y$)}
\end{subfigure}%
\hfill
\begin{subfigure}{0.31\linewidth}
\centering
\fbox{\includegraphics[width=\linewidth]{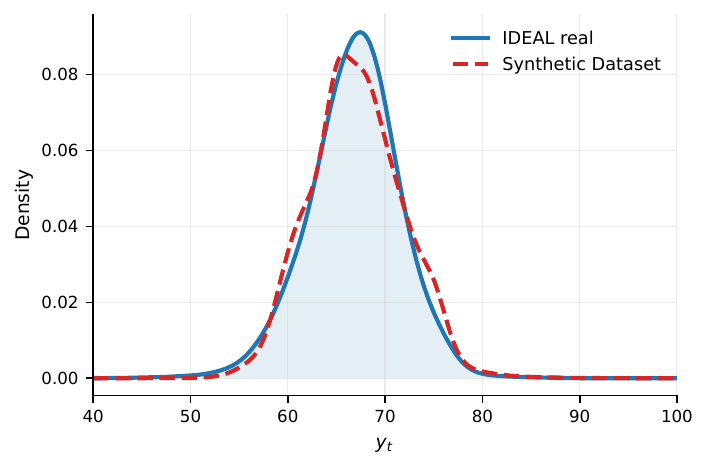}}
\caption{IDEAL ($y$)}
\end{subfigure}%
\hfill
\begin{subfigure}{0.31\linewidth}
\centering
\fbox{\includegraphics[width=\linewidth]{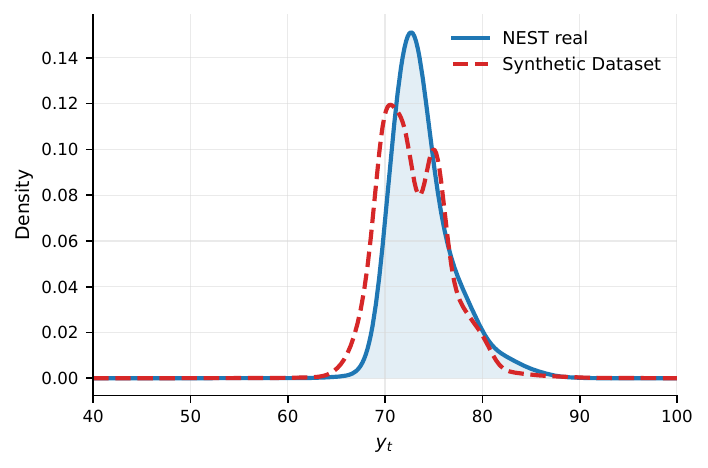}}
\caption{NEST ($y$)}
\end{subfigure}

\vspace{0.5em} % space between rows

% -------- Second row --------
\begin{subfigure}{0.31\linewidth}
\centering
\fbox{\includegraphics[width=\linewidth]{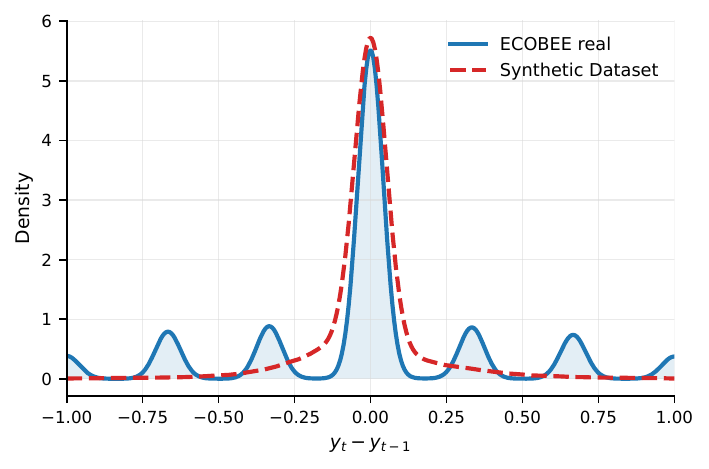}}
\caption{ECOBEE ($\Delta y$)}
\end{subfigure}%
\hfill
\begin{subfigure}{0.31\linewidth}
\centering
\fbox{\includegraphics[width=\linewidth]{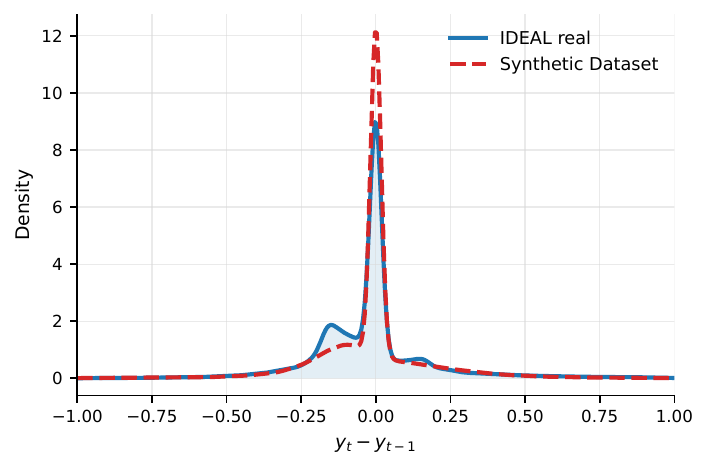}}
\caption{IDEAL ($\Delta y$)}
\end{subfigure}%
\hfill
\begin{subfigure}{0.31\linewidth}
\centering
\fbox{\includegraphics[width=\linewidth]{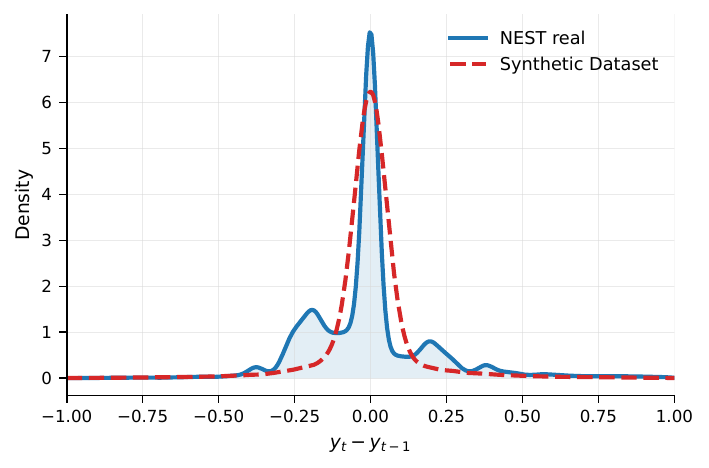}}
\caption{NEST ($\Delta y$)}
\end{subfigure}

\caption{Distributions of indoor temperature values ($y$) in the top row and their consecutive differences ($\Delta y$) in the bottom row across real and their corresponding synthetic datasets.}
\end{figure}

% ---------------------
% Total Load Distributions
% ---------------------
\begin{figure}[htbp]
\centering

% -------- First row --------
\begin{subfigure}{0.31\linewidth}
\centering
\fbox{\includegraphics[width=\linewidth]{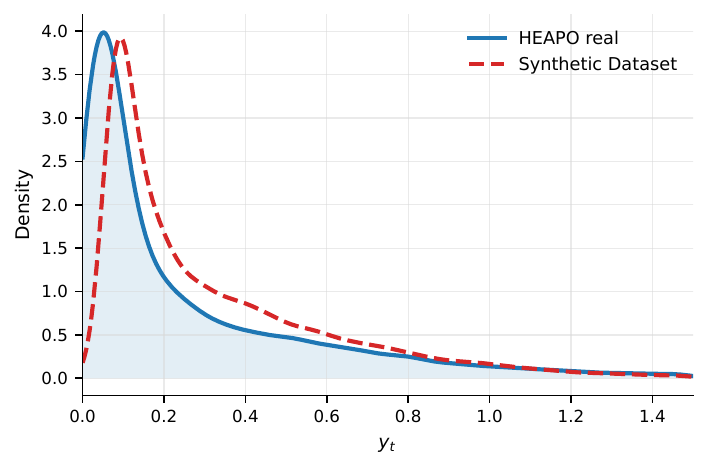}}
\caption{HEAPO ($y$)}
\end{subfigure}%
\hfill
\begin{subfigure}{0.31\linewidth}
\centering
\fbox{\includegraphics[width=\linewidth]{Figures/real_home_distribution_match/ideal/temperature_forecasting/distribution_comparison_x.pdf}}
\caption{IDEAL ($y$)}
\end{subfigure}%
\hfill
\begin{subfigure}{0.31\linewidth}
\centering
\fbox{\includegraphics[width=\linewidth]{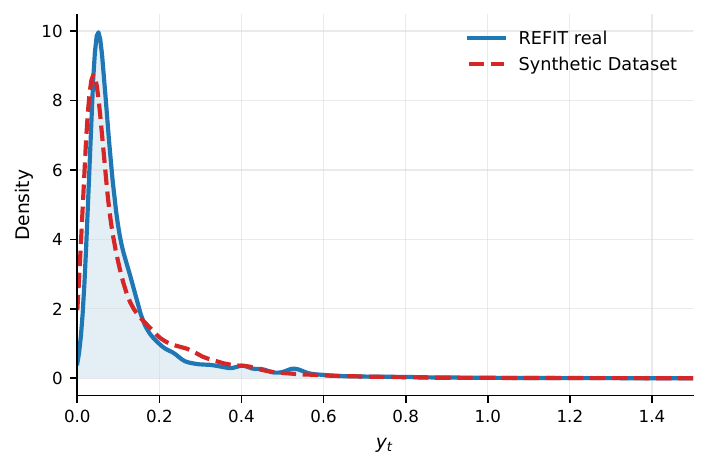}}
\caption{REFIT ($y$)}
\end{subfigure}

\vspace{0.5em}

% -------- Second row --------
\begin{subfigure}{0.31\linewidth}
\centering
\fbox{\includegraphics[width=\linewidth]{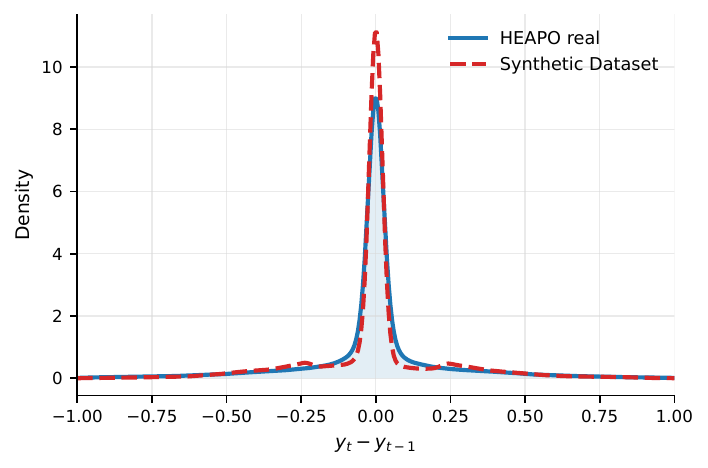}}
\caption{HEAPO ($\Delta y$)}
\end{subfigure}%
\hfill
\begin{subfigure}{0.31\linewidth}
\centering
\fbox{\includegraphics[width=\linewidth]{Figures/real_home_distribution_match/ideal/temperature_forecasting/distribution_comparison_delta.pdf}}
\caption{IDEAL ($\Delta y$)}
\end{subfigure}%
\hfill
\begin{subfigure}{0.31\linewidth}
\centering
\fbox{\includegraphics[width=\linewidth]{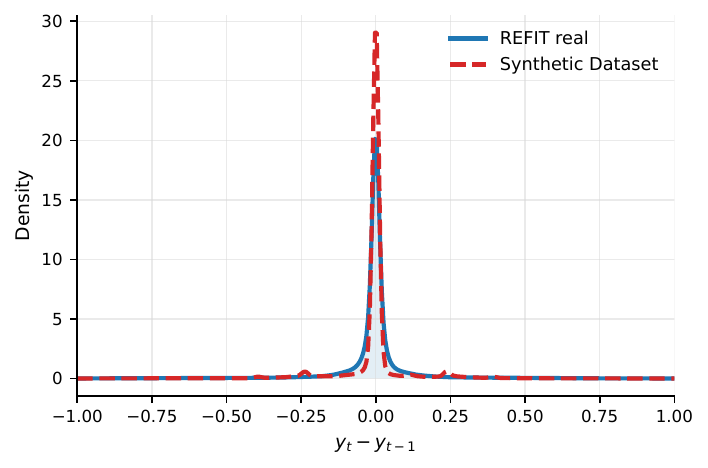}}
\caption{REFIT ($\Delta y$)}
\end{subfigure}

\caption{Distributions of total load values ($y$) in the top row and their consecutive differences ($\Delta y$) in the bottom row across real and their corresponding synthetic datasets.}
\end{figure}

%\newpage
%\input{checklist.tex}

\end{document}